\documentclass[a4paper,fleqn]{cas-sc}

\usepackage[numbers]{natbib}
\usepackage{subfigure}
\usepackage{soul,amsmath}

\def\tsc#1{\csdef{#1}{\textsc{\lowercase{#1}}\xspace}}
\tsc{WGM}
\tsc{QE}
\tsc{EP}
\tsc{PMS}
\tsc{BEC}
\tsc{DE}

\begin{document}
\let\WriteBookmarks\relax
\let\printorcid\relax
\def\floatpagepagefraction{1}
\def\textpagefraction{.001}
\shorttitle{A spatial-temporal short-term traffic flow prediction model based on Location-GCN and LSTM}
\shortauthors{Zhijun Chen et~al.}

\title [mode = title]{A spatial-temporal short-term traffic flow prediction model based on dynamical-learning graph convolution mechanism}   

	\tnotemark[1]
	
	\tnotetext[1]{This document is the results of the research project funded in part by the National Key R$\&$D Program of China [grant number 2018YFB1600600]; in part by the National Natural Science Foundation of China [grant numbers 52072288, 71702066]; in part by the Fundamental Research Funds for the Central Universities [grant number WUT: 2021VA004]; and in part by the Hubei Province Science and Technology Major Project[grant number 2020AAA001]. Yishi Zhang is the principal corresponding author of this paper.}                   
\author[1]{Zhijun Chen}[style=chinese]
\address[1]{Intelligent Transportation Systems Research Center, Wuhan University of Technology, Wuhan 430000, China}
\credit{Data curation, Formal analysis, Methodology, Resources, Writing {--} original draft, Funding acquisition}

\author[2]{Zhe Lu}[style=chinese]
\address[2]{School of Transportation and Logistics Engineering, Wuhan University of Technology, Wuhan 430000, China}
\credit{Data curation, Investigation, Writing {--} editing, Methodology, Software, Visualization}

\author[3]{Qiushi Chen}[style=chinese]
\address[3]{School of Computer Science and Technology, Wuhan University of Technology, Wuhan 430000, China}
\credit{Data curation, Writing {--} editing, Methodology, Software}

\author[3]{Hongliang Zhong}[style=chinese]
\credit{Data curation, Investigation, Methodology, Writing {--} original draft, Software}

\author[4]{Yishi Zhang}[style=chinese]
\cormark[1]
\ead{yszhang@whut.edu.cn }
\credit{Writing {--} editing, Conceptualization, Formal analysis, Methodology, Supervision, Funding acquisition}
\address[4]{School of Management, Wuhan University of Technology, Wuhan 430000, China}

\author[5]{Jie Xue}[style=chinese]
\address[5]{Faculty of Technology, Policy and Management, Safety and Security Science Group (S3G), Delft University of Technology, 2628BX Delft, The Netherlands}
\credit{Investigation, Resources, Software, Visualization}

\author[1]{Chaozhong Wu}[style=chinese]
\cormark[1]
\ead{wucz@whut.edu.cn}
\credit{Conceptualization, Formal analysis, Validation, Funding acquisition}

\cortext[cor1]{Corresponding author}

\begin{abstract}
Short-term traffic flow prediction is a vital branch of the Intelligent Traffic System (ITS) and plays an important role in traffic management. Graph convolution network (GCN) is widely used in traffic prediction models to better deal with the graphical structure data of road networks. However, the influence weights among different road sections are usually distinct in real life, and hard to be manually analyzed. Traditional GCN mechanism, relying on manually-set adjacency matrix, is unable to dynamically learn such spatial pattern during the training. To deal with this drawback, this paper proposes a novel location graph convolutional network (Location-GCN). Location-GCN solves this problem by adding a new learnable matrix into the GCN mechanism, using the absolute value of this matrix to represent the distinct influence levels among different nodes. Then, long short-term memory (LSTM) is employed in the proposed traffic prediction model. Moreover, Trigonometric function encoding is used in this study to enable the short-term input sequence to convey the long-term periodical information. Ultimately, the proposed model is compared with the baseline models and evaluated on two real word traffic flow datasets. The results show our model is more accurate and robust on both datasets than other representative traffic prediction models.
\end{abstract}


\maketitle

\section{Introduction}

As a crucial task in the Intelligent Traffic System (ITS) research, short-term traffic flow prediction has drawn more and more attention in recent years. A more accurate traffic prediction usually brings a better allocation of limited public transportation resources \citep{RN1,RN2,RN3}. Precise traffic prediction could also help the city administrators better manage the urban traffic, avoiding potential accidents, risk situations, or traffic paralysis beforehand.

The purpose of traffic flow prediction is to predict several road nodes' future traffic flow in an area based on the historical traffic data. The traffic flow in the same position usually has a steady temporal changing pattern, which means a strong correlation between the historical and future flow. Thus, traditional traffic prediction models like autoregressive integrated moving average (ARIMA) and linear regression (LR) conduct prediction based on the analysis of the temporal pattern of traffic flow \citep{RN4,RN5,RN6}.  \citet{RN7} consider the traffic flow as an evolution of variables, and propose a Lagrangian continuum traffic model, bridging microscopic and macroscopic approaches. However, those traditional mechanisms cannot efficiently deal with multi-feature sequence input, which is commonly seen in traffic prediction tasks, limiting their prediction accuracy. 

In recent years, artificial neural network (ANN) shows excellent performance in traffic prediction tasks due to its outstanding ability of pattern fitting and data processing. Inspired by the simple architecture and good performance in nonlinear relationship fitting of ANN, \citet{RN8} use multi-layer perceptron (MLP) and ARIMA to conduct traffic state prediction. Recurrent neural network (RNN) model, long short-term memory (LSTM) model and gated recurrent unit (GRU) model, are all widely-used ANN models which can analyze temporal pattern hidden in traffic data \citep{RN10,RN11,RN12,RN13}. Since the spatial pattern is also a vital factor that needs to be seriously considered during traffic prediction, there are also many widely-used spatial pattern analysis models. Convolutional neural network (CNN)  and graph convolutional network (GCN) are typical examples of such methods \citep{RN14,RN15,RN16,RN55}. Graph wavelet gated recurrent (GWGR) neural network \citep{RN17} uses wavelet transform to detect sudden changes and peaks in temporal signal, considering that traffic status on a road segment is highly influenced by the upstream/downstream segments and nearby bottlenecks. All of those models can take the mutual influences of nearby areas into consideration.

However, traffic flow changing is affected by both temporal factors and spatial factors. The most recent studies have been aware of this and are trying to develop spatial-temporal prediction methods. Among those studies, the RNN and GCN models usually are empolyed to compose the hybrid model, such as Diffusion Convolutional Recurrent Neural Network (DCRNN) proposed by \citet{RN18}. RNN can process the temporal traffic pattern properly due to its good ability of memory, meanwhile GCN can be better applied in real-life tasks than CNN, for it can deal with the graphical road network data. The classical GCN mechanism relies on the manually-set adjacency matrix to convey the information of the influence weights among different nodes. The elements of the adjacency matrix are usually set to the connectivity or distance of the road nodes to represent the different influence weights among different road nodes. However, the distinct influence degrees are usually hard to manually analyzed without training. When the connectivity is setted as the elements of the adjacency matrix, the adjacency matrix will become a 0-1 matrix, which means all the upstream nodes of a downstream node contribute the same influence weights in the convolution layer. This is completely against the real-life scenarios and will reduce the prediction accuracy. 

To better deal with those shortages, a spatial-temporal traffic flow prediction method called Location Graph Convolution Long Short Term Memory (Loc-GCLSTM) is proposed in this paper. It uses the absolute value of a newly-added trainable matrix of GCN to dynamically learn the different influence weights among nodes during the training progress. Moreover, Loc-GCLSTM attempts to place the periodic pattern of traffic data into the model inputs using Trigonometric Function Encoding. The main contributions of our work are listed as follows:

(a) Location-GCN is put forward to allow the convolution operation to dynamically learn the distinct influence weights among different road section nodes, and makes the GCN mechanism more coincident with the real-life scenario and improves the performance of the GCN model.

(b) A spatial-temporal traffic prediction model (Loc-GCLSTM) is proposed, which combines the proposed Location-GCN and LSTM to better capture the spatial-temporal information behind the real-life traffic data. 

(c) Experiments based on two real-world traffic network datasets are conducted to verify the effectiveness of the proposed method.

The remainder of this paper is organized as follows: Section 2 describes the related work and research status in the traffic prediction domain. Section 3 illustrates a specified description of the application scenario and the technical details of the proposed model. Then in Section 4, several experiments based on our OpenITS dataset and METR-LA dataset are designed and conducted to evaluate the proposed model comprehensively. Finally, Section 5 summarizes the work and raises some potential points which worth further study for improvement.

\section{Literature Review}

\subsection{Classical Traffic Prediction Methods}

Early models on traffic prediction such as auto-regression (AR), autoregressive moving average (ARMA), and ARIMA, conduct the task based on AR and data stationary assumption \citep{RN19,RN4,RN5}. However, these models can only do temporal analysis on long-time-span data and can only be fit and tested in a single same road, making them hard to be applied in the dynamic and volatile real-life traffic prediction. With the development of machine intelligence techniques, some studies try to apply machine learning to traffic prediction. For example, some of the classical machine learning methods like $k$-nearest neighbor ($k$NN), support vector regression (SVR), and extreme gradient boosting (XGBoost) have been applied to traffic flow prediction\citep{RN22,RN23,RN24}. To achieve better robustness in prediction, some researchers also use the combination of different machine learning models. \citet{RN25} combine XGBoost with the light gradient boosting machine (LightGBM) \citep{RN26} to conduct the traffic prediction task. \citet{RN27} build a model based on spatio-temporal correlations using a multivariate spatial-temporal autoregressive (MSTAR) model. \citet{RN28} use the stochastic differential equations (SDEs) to optimize a two-step prediction based on the Hull-White (HW) model and the Vasicek model (EV).

\subsection{Traffic Prediction using Single Deep Learning Model}

In recent years, a series of research try to use ANN, such as MLP and AutoEncoder, to predict traffic flow\citep{RN9,RN29,RN53}. RNN is firstly used to analyze the changing patterns from historical traffic data for the temporal models. However, RNN is not able to analyze long-term temporal dependencies, which usually last for quite a number of moments. To this end, the LSTM and the GRU are put forward recently. To better analyze the bi-directional temporal pattern in flow changing, \citet{RN35} apply Bi-LSTM to generate the traffic pattern from both original and reversed order data using Bi-directional LSTM, achieving higher accuracy in prediction tasks. To better deal with lane-level traffic data, \citet{RN34} use entropy-based grey relation analysis (EGRA) to analyze the influence among different lanes, and present a lane-level traffic prediction model combining GRU and LSTM. Nested LSTM puts an LSTM unit inside an LSTM to improve the learning process's stability, hoping to make the learning of long-term traffic patterns in historical traffic data more stable \citep{RN36,RN37}. Inspired by natural language processing (NLP) tasks, many researchers also try to utilize the Sequence to Sequence (Seq2Seq) and attention mechanism in traffic prediction, expecting to obtain more accurate prediction results \citep{RN38,RN39}. 

Spatial information is also of great value in traffic flow prediction. When it comes to the spatial models, the convolution operation is usually applied to generalize and analyze the spatial correlations among nearby locations or road sections. CNN is firstly used in traffic flow prediction to analyze several nearby locations' mutual influences in a transportation network \citep{RN14}. Because CNN cannot efficiently analyze Non-Euclidean structure data , GCN is then introduced into traffic prediction to deal with graph-structured data, such as a transportation network composed of various road section nodes \citep{RN16}. Since the road network in GCN is not able to dynamically change over time , Graph attention (GAT)  uses the attention mechanism to describe the correlations among different road section nodes and analyze graph structure traffic data \citep{RN40,RN41}.\citet{RN42} develop Multi-GCN model which introduces many other spatial factors among nodes such as a distant map into the normal GCN mechanism for better awareness of various traffic information . 

\subsection{ Spatial-temporal Traffic Prediction}

However, real-life traffic flow is influenced by both spatial and temporal factors. Recent research starts to focus on analyzing traffic patterns from both spatial and temporal perspectives \citep{RN18,RN57,RN56}. DCRNN \citep{RN18}, temporal graph convolutional network (T-GCN) \citep{RN43}, and sequence to sequence model based on graph convolution (GC-Seq2Seq) \citep{RN44} are the three representative methods that directly combine spatial layers with temporal layers. To better use the mutual influence between spatial and temporal factors, \citet{RN45} utilize spatial-temporal blocks for data analysis from both spatial and temporal dimensions, and propose spatial-temporal graph convolutional network (STGCN) model.In STGCN model, the temporal convolutional layers are realized by CNN, and the graph convolutional layers are realized by GCN, trying to better analyze the correlations between nearby moments and locations. To more flexibly deal with the dynamical changing traffic network, \citet{RN46} propose Graph Attention LSTM Network (GAT-LSTM) to predict traffic flow ,which uses GAT to update spatial information and uses LSTM to generalize the temporal patterns hidden in historical traffic data in such a way as to conduct spatial-temporal analysis dynamically. Attention graph convolutional sequence-to-sequence model (AGC-Seq2Seq) is proposed by \citet{RN47} to train Seq2Seq model in traffic prediction, also using GCN and attention mechanism to promote accuracy. \citet{RN54} similarly consider the spatio-temporal feature of traffic data and propose a dynamic graph recurrent convolutional neural network for traffic flow prediction.

Overall, spatial-temporal deep learning models are the main trend of recent short-term prediction research. Although those studies can achieve a higher prediction accuracy ratio than traditional regression models, most of them still have some shortages. Firstly, GCN can't analyze and distinguish various upstream nodes' different influence weights toward the same downstream node as GAT, although GCN can usually perform better than GAT due to its superior learning ability. Furthermore, as ANN models usually take a not-long sequence as input, it is hard to convey periodic information across weeks or even months in those short-term prediction models. Besides, due to the better performance and stronger generalization ability of GCN and LSTM, this paper tries to promote the above drawbacks based on an improved GCN-LSTM hybrid model. 

\section{Methodology}

\subsection{Problem Description}

The proposed ANN model aimed at better dealing with the short-term traffic flow prediction tasks. In such tasks, history traffic data of several positions in an area is used to predict the future traffic data of those same positions.

In this study, the digraph $ G=(V,E,A) $ is used to describe the urban road network area where our traffic prediction task is conducted, where $ V $ is the set of road section nodes, $ E $ denotes the set of edges among road section nodes, and $ A $ represents the adjacency matrix of section nodes in the target area. $ A_{i,j} $ means whether the ith section is connected with jth section ($ A_{i,j} $=1 if they are connected and 0 otherwise). The values of the diagonal in the adjacency matrix are all 0. When it comes to an area with $ N $ road section nodes, the corresponding adjacency matrix can be denoted as $ A \in R^{N \times N} $. In this study, different directions of a road are considered as two different road sections.

\subsection{Spatial-Temporal Analysis}

Traditional traffic flow prediction models, no matter GRU, LSTM, GCN, or GAT, concentrating on analyzing the traffic pattern from a single perspective. However, the real-life traffic flow changing is affected by both spatial and temporal factors. To better analyze the historical traffic data and achieve a more accurate prediction performance, the historical traffic data need to be investigated from both spatial and temporal perspectives.

Based on thorough consideration, the LSTM and GCN are chosen as the basic methods in our model to conduct spatial and temporal analysis.

\subsubsection{Long Short-Term Memory}

\begin{figure}[htb]
	\centering
	\includegraphics[scale=.5]{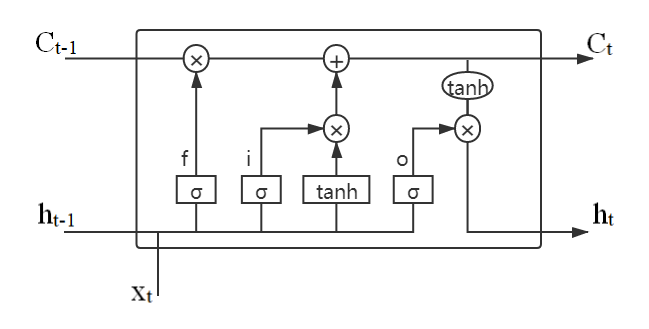}
	\caption{Unit structure of LSTM.}
	\label{FIG:1}
\end{figure}

Among all temporal analysis mechanisms in traffic flow prediction, LSTM is the most efficient, widely-used, and representative one. So it is chosen to be used in our model for temporal pattern processing. LSTM uses cell state and three gated structures: the input gate $ i $, the forget gate $ f $, and the output gate $ o $, to realize the analysis, memory, and output of the temporal sequences’ long-term dependencies. Figure \ref{FIG:1} shows the structure of an LSTM unit. 

\begin{align}\label{eq1}
f_t & =\sigma(V_f h_{t-1}+W_f X_t+b_f)\\
i_t & =\sigma(V_i h_{t-1}+W_i X_t+b_i)\\
\tilde{C}_t &=\tanh (V_C h_{t-1}+W_C X_t+b_C)\\
C_t & =f_t \cdot C_{t-1}+i_t \cdot \tilde{C}_t\\
o_t & =\sigma(V_o h_{t-1}+W_o X_t+b_o)\\
h_t & =o_t \cdot \tanh C_t
\end{align}

The above equations describe the whole processing flow of LSTM, where $ \sigma $ and $ tanh $ are activation functions, Equation 1 indicates the process of forget gate, Equation 2 and 3 indicate the process of input gate, Equation 4 represents the fusion process of information from the above two  processes, Equation 5 and 6 represent the process of output gate.

\subsubsection{Graph Convolution Network}

GCN can efficiently deal with graph-structured data such as urban traffic network data. It is used in our model for spatial analysis. When processing graph-structured data with $ N $ nodes the traditional GCN mechanism in a digraph road network can be written as:

\begin{eqnarray}\label{eq2}
H_l=
\begin{cases}
X, &l = 0\\
(A+E) H_{l-1} W_l &l\ge 1
\end{cases}
\end{eqnarray}

\begin{figure}[htb]
	\centering
	\includegraphics[scale=.3]{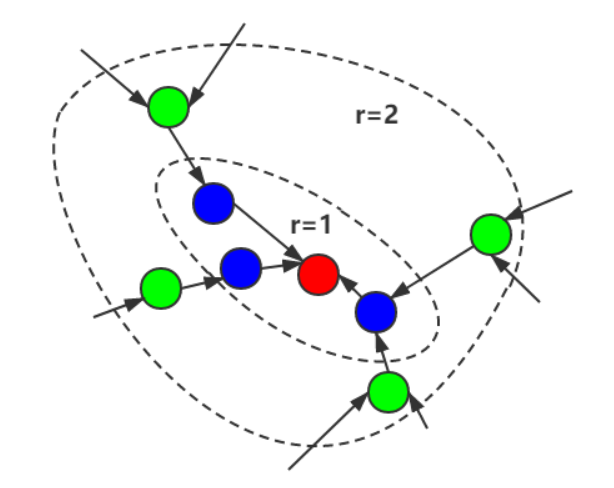}
	\caption{Graph convolutional algorithm spread layer by layer.}
	\label{FIG:2}
\end{figure}

$ A \in R^{N \times N} $ is the adjacency matrix, $ X \in R^{N \times F} $ denotes the $ F $ features’ input data of $ N $ nodes, $ H_l \in R^{N \times F'} (l \ge 1) $ means the convolution result of $ F' $ features after $ lth $ convolution, $ E $ is the identity matrix and $ W $ presents the trainable matrix. Figure \ref{FIG:2} shows the physical meaning of the graph convolution. The convolution operation is a weighted summation to several nearby data in one area. When convoluting the red target nodes in the picture, if $ l=1 $, the algorithm performs a weighted summation of the traffic data of target nodes and their upstream nodes. Then it updates the target nodes’ data, integrating upstream flow information into it; if $ l=2 $, the algorithm performs a weighted summation of the green points whose distance are 2, then updates the target nodes’ data, spreading layer by layer until reaching the manually set maximum distance limit.

The traditional GCN mechanism uses $ A+E $ to describe graph information, where the adjacency matrix $ A $ and the identity matrix $ E $ consist of only 0 and 1. The target node with more than 1 upstream nodes’ data gets bigger and bigger during the graph convolution. Therefore, the inverse matrix of the digraph’s degree matrix is usually used to multiply the matrix $ A+E $, normalizing data after every convolution operation. The equation can be written as:

\begin{eqnarray}\label{eq3}
H_l=
\begin{cases}
X, &l = 0\\
D^{-1}(A+E) H_{l-1} W_l &l\ge 1
\end{cases}
\end{eqnarray}

\subsection{Location Graph Convolution Network}

\subsubsection{Trainable Matrix}

\begin{figure}[htb]
	\centering
	\includegraphics[scale=.5]{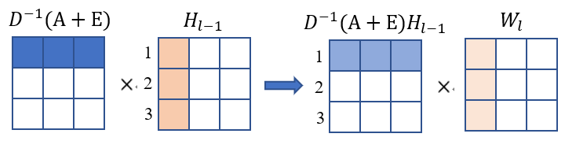}
	\caption{Drawbacks of traditional Graph Convolution.}
	\label{FIG:3}
\end{figure}

However, the classical graph convolution mechanism still has some drawbacks. As shown in Figure \ref{FIG:3}, every row in matrix $ H_{l-1} $ means the traffic data of a single node in the road network. During its multiply with $ D^{-1}(A+E) $, elements in a different row of $ H_{l-1} $ are weighted summed to form a new row, which means the traffic data in different upstream nodes are conveyed into a single downstream node, updating the data of the target nodes. However, in the multiply between this weighted sum result and the original trainable matrix $ W_l $, there is no interaction between the two different rows, or to say, nodes. So in fact, the only matrix involved in the weighted sum between upstream and downstream nodes, aside from the node traffic data matrix $ H_{l-1} $ itself, it’s the $ D^{-1}(A+E) $. However, all elements in $ D^{-1}(A+E) $, which represent the influence weights between the two different nodes, are manually set before the training. And in most cases, the different influence relationship among various road section nodes is hard to be analyzed manually. It’s usually better to learn that knowledge by model training. Moreover, in most cases, the values of the adjacency matrix are set to 0 and 1, which means the mechanism tacitly approves that every different upstream nodes’ influence on the target downstream node is the same, totally against reality. Both points mentioned above are negative for the model to better analyse the data’s traffic pattern and achieve high accuracy in prediction.

To overcome this drawback, since the original trainable matrix $ W_l $ cannot be used for dynamical learning, a trainable matrix $ W_{mask} $ is put forward to learn the different weights between the two road nodes. By using the $ W_{mask} $, we also look forward to enabling the improved GCN the ability to automatically adjust the influence weights during the training process.

$ W_{mask} $ is a $ N \times N $ matrix where $ N $ is the number of road section nodes. It's in the same shape as the adjacency matrix $ A $. Value at position $ (i,j) $ means the weight of $ ith $ section’s influence on the $ jth $ section. Compared with the traditional GCN mechanism where $ A+E $ is used to take the graph information between nodes into the analysis, this matrix can make different node pairs’ influences independently changing via the training, but not just be set as 0 or 1 compared with the traditional GCN mechanism. This initial weights of this matrix can be randomly set, but not recommended to set as all 0 since it does not correspond with real life. Through this, the information on the difference in influence levels among different nodes can be learned.

\subsubsection{Calculating Improvement}

To make the proposed GCN mechanism work, the calculation method of GCN needs to be changed. After the pattern learning of $ W_{mask} $, the absolute value of each element in $ W_{mask} $ is used to make a new matrix $ W_{abs} $, for the influence of upstream roads to downstream roads are usually positive. It is then used to dot multiply the $ A+E $ matrix, aiming at using the different influence levels among various nodes learned from real-life training data to adjust the values of the adjacency matrix. Thus, enable the model to learn the different influence levels among various road nodes. The improved mechanism can be described by the equations below:

\begin{align}\label{eq4}
W_{abs} &= \vert W_{mask} [i,j] \vert \quad(i\in [0,N-1],j\in [0,N-1])\\
H_l &= 
\begin{cases}
X, &l = 0\\
D^{-1}(W_{abs} \cdot (A+E)) H_{l-1} W_l &l\ge 1
\end{cases}
\end{align}

\begin{figure}[htb]
	\centering
	\includegraphics[scale=.5]{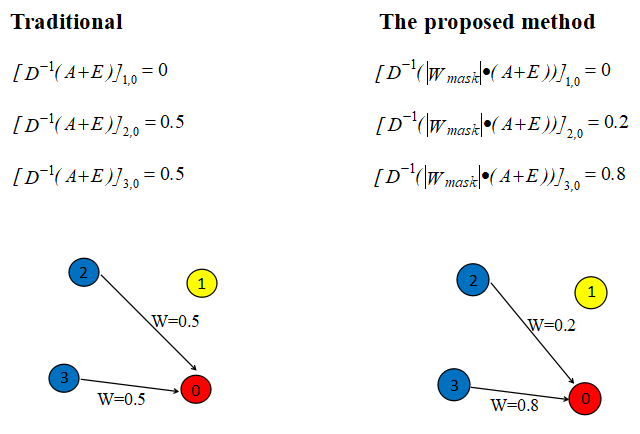}
	\caption{Drawbacks of traditional Graph Convolution.}
	\label{FIG:4}
\end{figure}

By using $ W_{mask} $ and $ W_{abs} $, Location-GCN can look into a more detailed spatial pattern in traffic flow. A comparison sample is shown in Figure \ref{FIG:4} between traditional GCN and the proposed method.

In a road network with 4 nodes, where node 2 and node 3 are the upstream nodes of node 0. Since traditional GCN can only represent the influence levels between point pairs with value 0 or value 1, the weight value at the position $ [2,0] $ and $ [3,0] $ in $ A+E $ are both 0.5 after the normalization of $ D^{-1} $, leading to the mechanism considers the influence weights of node 2 and node 3 toward node 0 are equal. However, our mechanism can enable the model to learn the difference in influences during the training and allow the convolution operation to get this knowledge via the dot multiply with $ W_{abs} $.

\subsection{Trigonometric Function Encoding}

Since for short-term traffic flow prediction, each input sample usually varies in a short time span like one or two hours. The periodic changing pattern among days or weeks cannot be directly analyzed by the neural network model in each sample's processing. Thus periodic data is needed to be encoded into the input samples of the network model.

Sine function and Cosine function, as periodic functions, can describe periodic information in traffic data. Thus, we use those trigonometric functions to encode the hour and moment information in the period. The time interval of the task is set to 5 minutes, so the time point data is first transferred to $ (i,j) $, meaning it is the $ ith $ 5 minutes in the day and the $ jth $ hour in a week. Then the Trigonometric Function Encoding encodes $ (i,j) $ into 4 columns $ moment\_sin $, $ moment\_cos $, $ hour\_sin $, $ hour\_cos $, using the following equations:

\begin{align}\label{eq5}
moment\_sin & = \sin \frac{2\pi i}{moment\_num}\\
moment\_cos & = \cos \frac{2\pi i}{moment\_num}\\
hour\_sin & = \sin \frac{2\pi j}{hour\_num}\\
hour\_cos & = \cos \frac{2\pi j}{hour\_num}
\end{align}

Where $ moment\_num $  is the total number of 5-minute-interval moments in a day, and $ hour\_num $ is the total number of hours in a week. Using this method, our prediction method is able to analyze the long-term periodic pattern in short-term traffic flow prediction tasks.

\subsection{The proposed model Loc-GCLSTM}

\begin{figure}[htb]
	\centering
	\includegraphics[scale=.5]{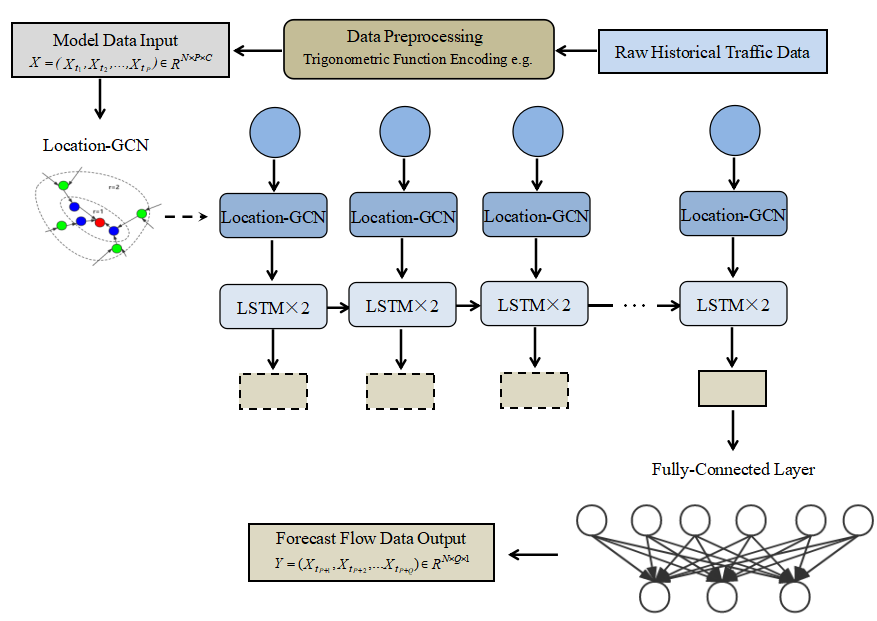}
	\caption{Structure of Loc-GCLSTM.}
	\label{FIG:5}
\end{figure}

To better deal with the periodic pattern in the historical traffic data, Loc-GCLSTM firstly apply Trigonometric Function to encode moment data, using the periodic property of the sine function and the cosine function to describe periodic information in the traffic pattern. Then the data is put into a Location-GCN layer, where a trainable matrix is utilized to learn various upstream road sections' different influence levels towards each downstream section. Ultimately, we combine the spatial-analyzing Location-GCN with temporal-analyzing LSTM to conduct prediction. After the spatial analysis conducted by Location-GCN and temporal analysis conducted by LSTM, the last step output of LSTM is chosen to put into a Fully-Connected Layer to generate the output forecast flow sequences, considering the last step output of LSTM is the semantic vector which generalized the pattern of sample flow. The structure of our model is shown in Figure \ref{FIG:5}. It is easy to know that the computational complexity of the spatial analysis is $\mathcal{O}(n^{2})$, where $n$ denotes input data size \citep{yu2017}. As for the temporal analysis, its computational complexity per time step is $\mathcal{O}(W)$, where $W=KH+KCS+CSI+HI$ denotes the number of weights, $K$, $C$, $S$, $H$ and $I$ denote the number of output units, the number of memory cell blocks, the size of the memory cell blocks, the number of hidden units, and the (maximal) number of units forward connected to memory cells, gate units, and hidden units, respectively \citep{hochreiter1997}. Hence the computationl complexity of Loc-GCLSTM per time step is $\mathcal{O}(n^2+W)$.

\section{Experiments and Results}

\subsection{Datasets}

\begin{figure}[htb]
	\centering
	\includegraphics[scale=.5]{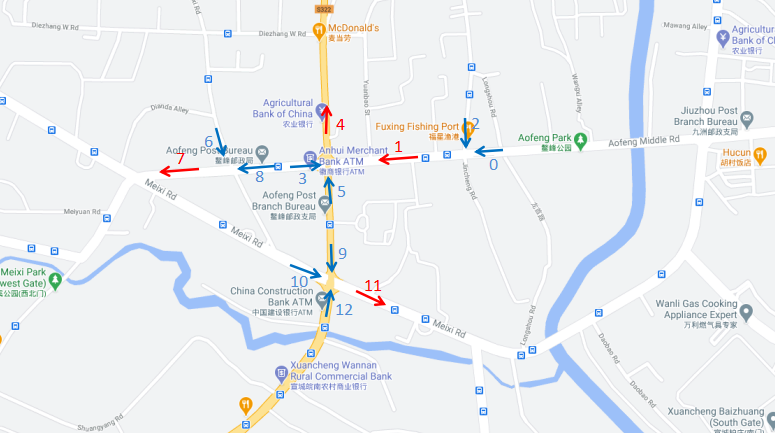}
	\caption{Locations of observation points in OpenITS dataset.}
	\label{FIG:6}
\end{figure}

We conduct experiments on two real-word datasets: OpenITS dataset and METR-LA\footnote{The raw dataset is available at https://github.com/liyaguang/DCRNN/tree/master/data} dataset. The \textbf{OpenITS} dataset is collected from the OpenITS platform, with data recorded between 3 o'clock to 24 o'clock each day from 1st July to 31st July (2nd July not included) in XuanCheng, Anhui Province, PRC. 13 road section observation nodes are chosen for experiments, and the time interval is set to 5 minutes. The specified locations of the observation nodes are shown in Figure \ref{FIG:6}. Every observation section is denoted by an arrow, and the direction of the arrow indicates the flow direction of the observation node. The red nodes are the selected test nodes. The obtained data is saved in the form of CSV files, and a new column of weather is added to the data manually. The test nodes' 06:30-09:30, 10:30-13:30, 17:00-22:00 data in 3rd, 5th, 7th, 15th, 16th, 18th, 27th July is chosen as the test dataset. All observation points' 03:00-24:00 data in the days other than 7 test days are set as the training dataset. In this study, the K-Nearest mechanism is applied to fill the missing data.

\begin{figure}[htb]
	\centering
	\includegraphics[scale=.5]{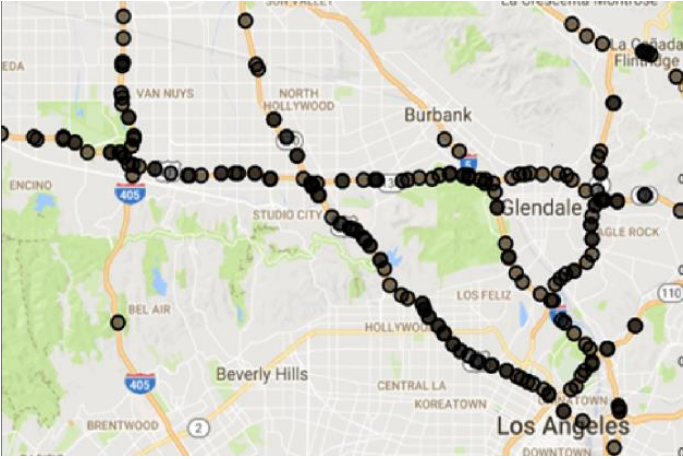}
	\caption{Locations of observation points in METR-LA dataset \citep{RN18}.}
	\label{FIG:7}
\end{figure}

\textbf{METR-LA} contains the all-day traffic data collected in 207 positions in the highway of Los Angeles County, from March 2012 to June 2012. We select the data in March and April as our experiment dataset. With more road sections, and a much more complicated traffic situation in Los Angeles compared with XuanCheng in OpenITS dataset, METR-LA can more evidently show the effectiveness of the promotions used in Loc-GCLSTM. The METR-LA dataset is randomly divided into training and test datasets with a proportion of four to one. All the data in every road section and moment is included in this dataset, which makes this experiment more challenging. The K-Nearest mechanism is also applied to fill the missing data of METR-LA dataset. Figure \ref{FIG:7} shows the observation point settings in METR-LA dataset.

\subsection{Data Preprocess}

The two datasets are preprocessed using the same method. Firstly, a sliding window is used to process the datasets: from the first 5 minutes of each day, generating samples every 2 consecutive hours (24-timesteps with 5-min interval), note that the 24-timesteps shall not be interrupted. Ultimately, 5440 samples are collected for OpenITS dataset and 17520 samples for METR-LA dataset. We apply five-fold cross-validation to conduct the experiments, namely, we randomly divide the dataset into five folds, using the first (second, third, fourth, and fifth) fold as the test set and the remaining folds as the training set, and finally reporting the average result as the performance indicator of each method used in our experiments.

The training and test samples are arranged into two NumPy arrays in the shape of $ [sample\_num,node\_num, $ $time\_lags,feature\_nums]$, where $ sample\_num $ denotes the total number of samples, $ node\_num $ presents the number of nodes in the road network, $ time\_lags $ means the timesteps (12 in our experiments) of each sample, $ feature\_nums $ is the number of features. The traffic flow, average speed, flow density, the proportion of big and regular vehicles, moment (5 minutes interval, 1-day period), hour (1-hour interval, 1-week period), lane number, and weather of every observation point are chosen as the input features in OpenITS dataset. Traffic flow, moment (5 minutes interval, 1-day period), hour (1-hour interval, 1-week period) are selected in METR-LA dataset. The time point and hour data are encoded by trigonometric function, while the weather is encoded by number. Those data are put into the model after z-score standardization.

\subsection{Evaluation Criteria and Comparison Models}

 Root mean square error (RMSE), mean absolute percentage error (MAPE), mean absolute error (MAE), median absolute percentage error (MdAPE), and median absolute error (MdAE) are chosen as the evaluation criteria in our experiments. These five criteria are all widely used in traffic prediction tasks. Note that MdAPE and MdAE take the median of the errors instead of their mean (in contrast to MAPE and MAE):

\begin{align}\label{eq6}
RMSE&=\sqrt{\frac{1}{n}\sum_{i=T+1}^{T+n} (\hat{Y}_i-Y_i)^2}\\
MAE&=\frac{1}{n}\sum_{i=T+1}^{T+n} \vert \hat{Y}_i-Y_i \vert\\
MAPE&=\frac{1}{n}\sum_{i=T+1}^{T+n} \vert \frac{\hat{Y}_i-Y_i}{Y_i} \times 100 \vert\\
MdAE&=median(\vert \hat{Y}_1-Y_1 \vert,\cdots,\vert \hat{Y}_n-Y_n \vert) \\
MdAPE&=median(\vert \frac{\hat{Y}_1-Y_1}{Y_1} \times 100 \vert,\cdots,\vert \frac{\hat{Y}_n-Y_n}{Y_n} \times 100 \vert)
\end{align}

In our experiments, three conventional machine learning methods, namely LR, XGBoost, and SVR, and three deep learning methods, namely LSTM, STGCN and DCRNN are selected as the comparison models. LR, XGBoost, SVR, and LSTM are temporal analysis models; STGCN and DCRNN are spatial-temporal prediction models. Detailed information on these models is given as follows.

\begin{itemize} 
\item LR \citep{RN6} is the most basic short-term traffic prediction method, which uses linear parameters to fit the training samples' pattern simply. LR usually uses ordinary least squares or linear least squares to update its parameters. After the data fitting, the LR model can be described as the below equation:

\begin{equation}\label{eq7}
Y=w_1 x_1+w_2 x_2+\cdots+w_n x_n
\end{equation}

\item XGBoost \citep{RN24} is an improvement based on gradient boosting decision tree (GBDT) \citep{RN48}, as it is the best performed and widest used Boosting model in recent traffic prediction competitions. XGBoost is an ensemble model based on several single tree models. Thus can reduce the sensitivity to the abnormal samples of the model, enhancing stability.

\item SVR \citep{RN22,RN23} is an outstanding machine learning mechanism focusing on vital samples, avoiding local data perturbation. It uses a soft margin mechanism to allow the samples to fluctuate in a small error range, improving the model's generalization ability. Aside from this, it selects vital samples as support vectors to conduct the data fitting, thus getting a better view of the global data samples, leading to a comparable high accuracy.

\item LSTM \citep{RN11,RN12}, as mentioned above, is the most classical neural network model used in the temporal analysis. With the use of cell state and Gated mechanism, LSTM can efficiently learn the long-term dependency pattern compared with other methods, making it commonly used in temporal analysis tasks. However, a single LSTM network cannot deal with the spatial factors, which are also crucial in traffic prediction.

\item STGCN \citep{RN45} combines temporal analysis using CNN and spatial analysis using GCN and introduces gated linear units (GLU) to optimize its output ability. CNN is used to generalize the temporal changing pattern of traffic data layer by layer. GLU is used to optimize the CNN output by enabling it to fit a more complex pattern. GCN is always put between every two CNNs, aiming at analyzing the spatial pattern of traffic flow. Using such a structure, STGCN is a spatial-temporal traffic prediction model that can take the mutual influence between two perspectives into consideration. However, since STGCN doesn't use the RNN-based mechanism for spatial analysis, it cannot thoroughly look into the flow changing details.

\item DCRNN \citep{RN18} is a spatial-temporal model which consists of bidirectional GCN and GRU. It puts a pair of trainable weights to separate learn the influence from both upstream and downstream positions of one node. The weighted sum of the upstream and downstream convolution will be further put into a multi-layer GRU network to further analyze the temporal changing pattern of the history traffic data. 

\end{itemize}

\subsection{Experimental Setting}
All experiments are compiled and tested on a Windows platform (CPU: Intel(R) Core(TM) i7-8700K CPU @ 3.70GHz, GPU: NVIDIA GeForce RTX 2080 Ti). Tensorflow is used by the implementation of all the neural network based models\citep{RN58}. During the training of the neural network, mean square error (MSE) is used as the loss function, while root mean square propagation (RMSProp) \citep{RN49,RN50,RN51} is used as the optimizer. The neural network model is trained for 600 epochs in OpenITS experiment and 300 epochs in METR-LA experiment. 

\begin{figure}[htb]
	\centering
	\begin{minipage}[]{0.45\linewidth}
		\centerline{\includegraphics[scale=.35]{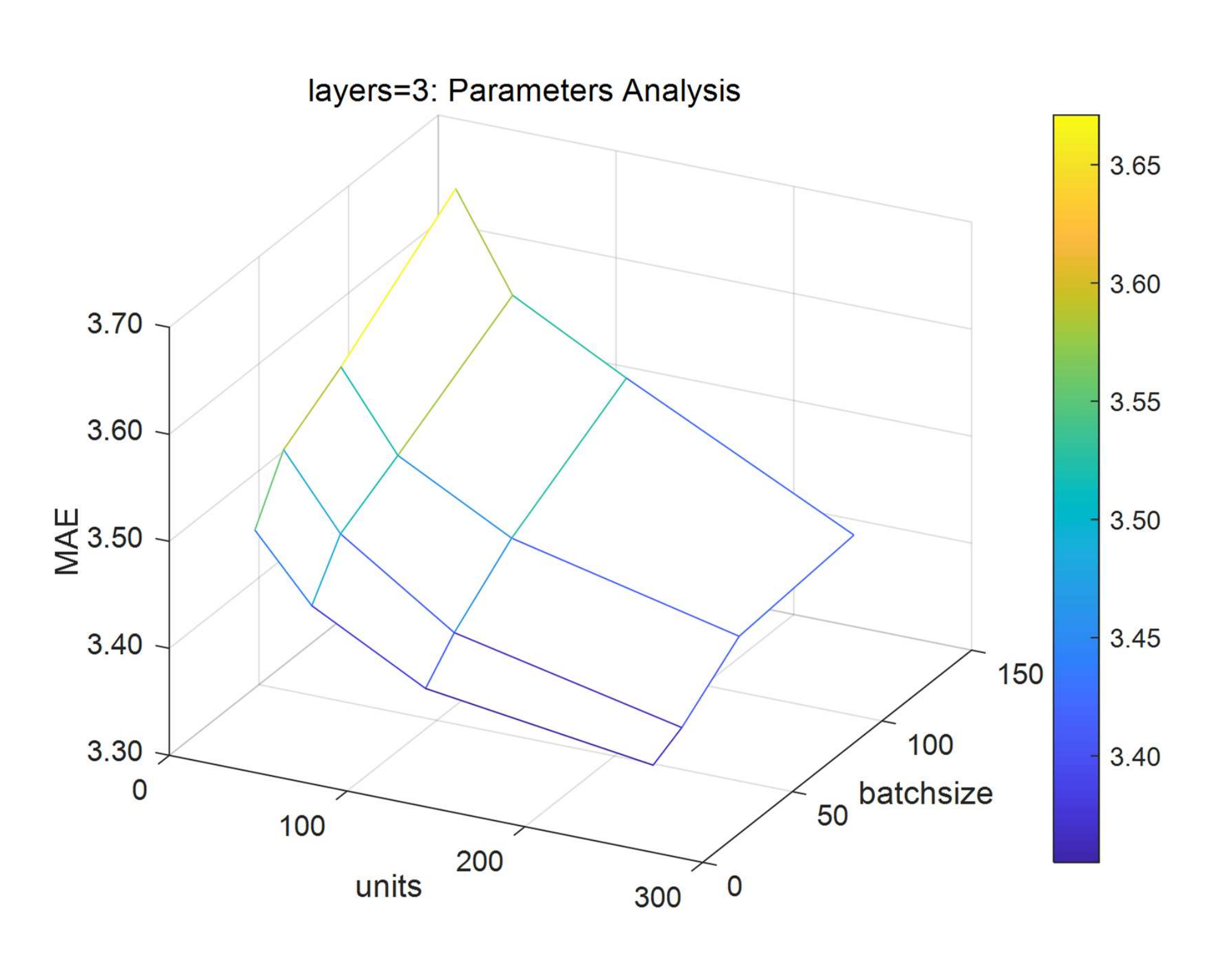}}
		\centerline{(a)}
	\end{minipage}  
	\begin{minipage}[]{0.45\linewidth}
		\centerline{\includegraphics[scale=.35]{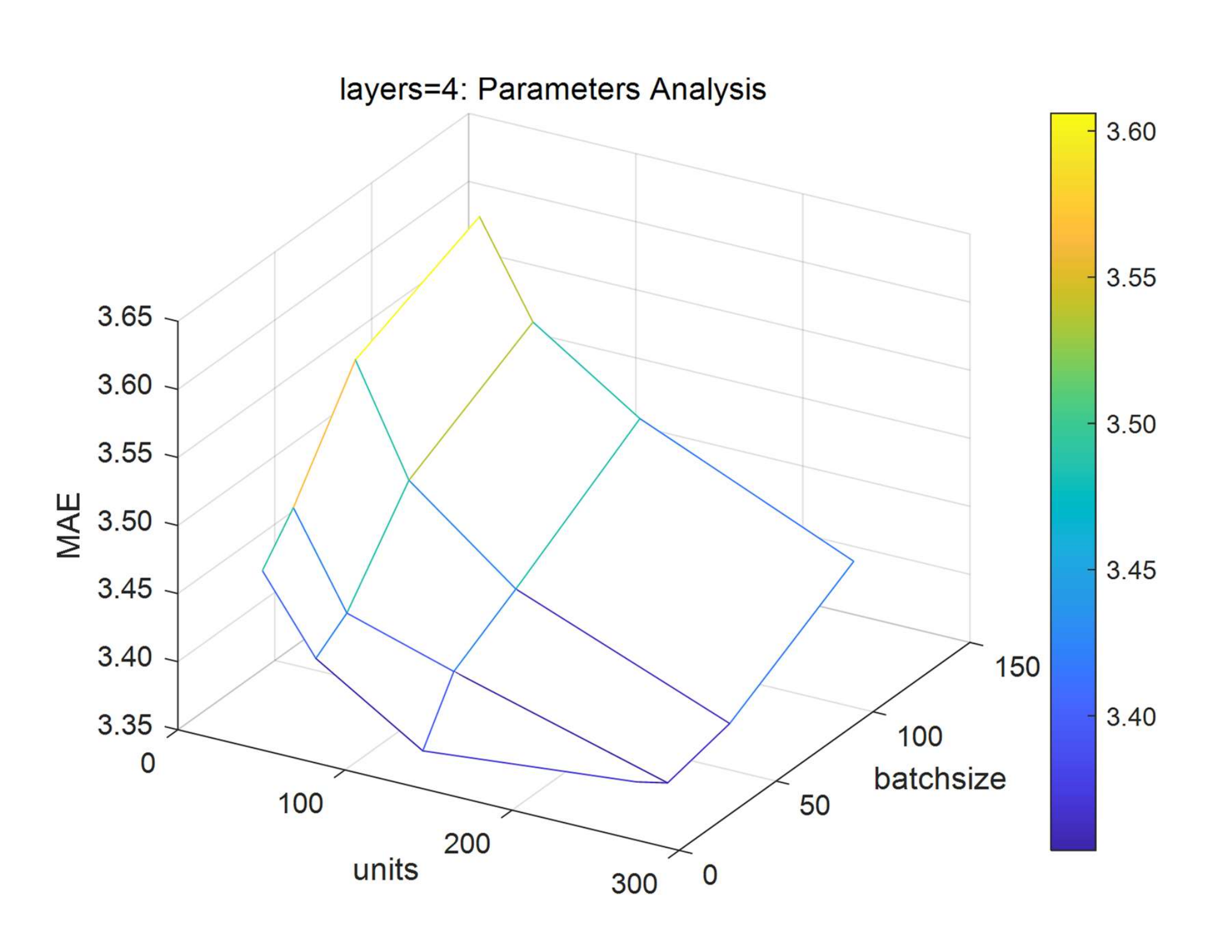}}
		\centerline{(b)}
	\end{minipage} 
	\begin{minipage}[]{0.45\linewidth}
		\centerline{\includegraphics[scale=.35]{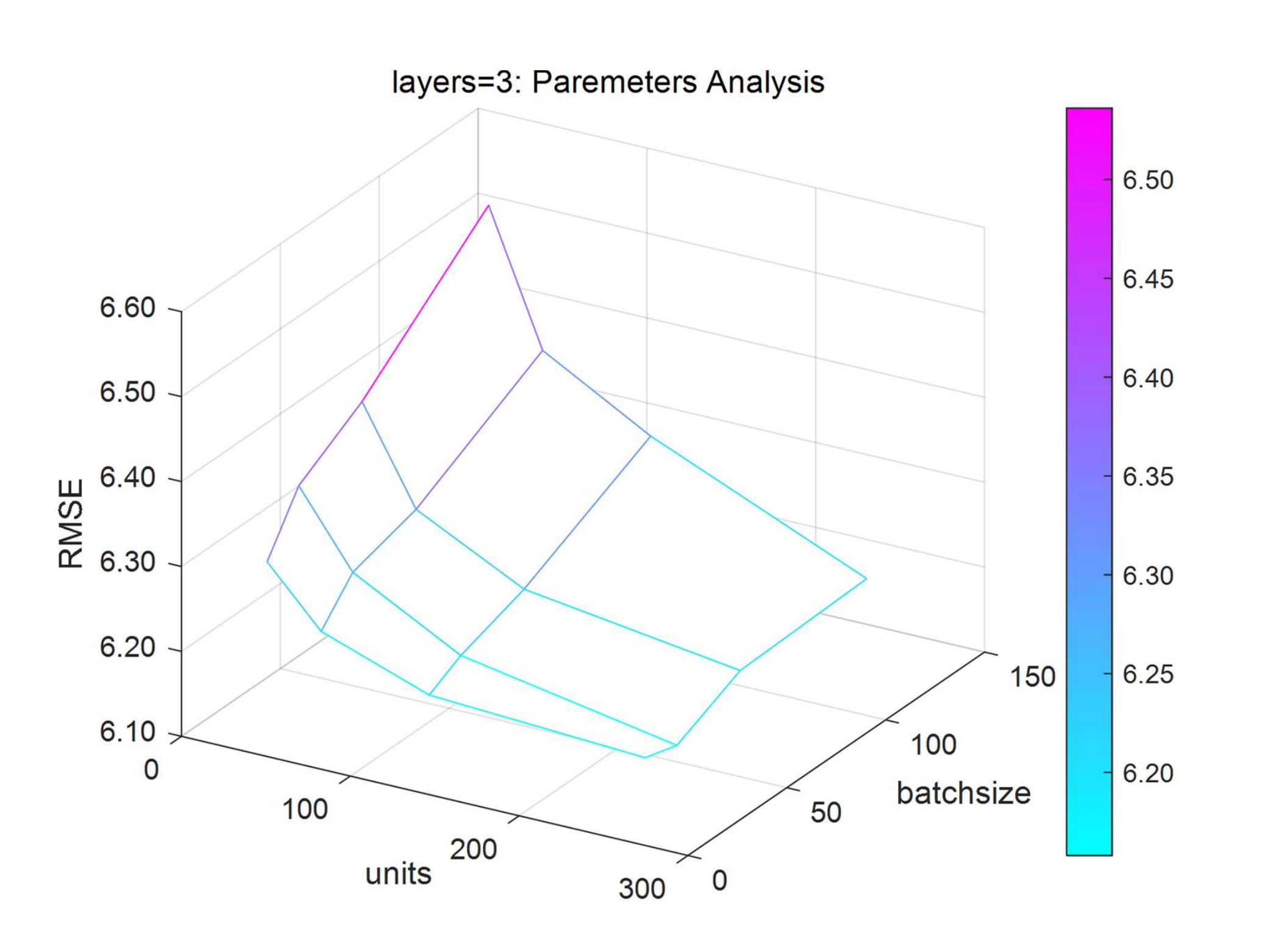}}
		\centerline{(c)}
	\end{minipage}  
	\begin{minipage}[]{0.45\linewidth}
		\centerline{\includegraphics[scale=.35]{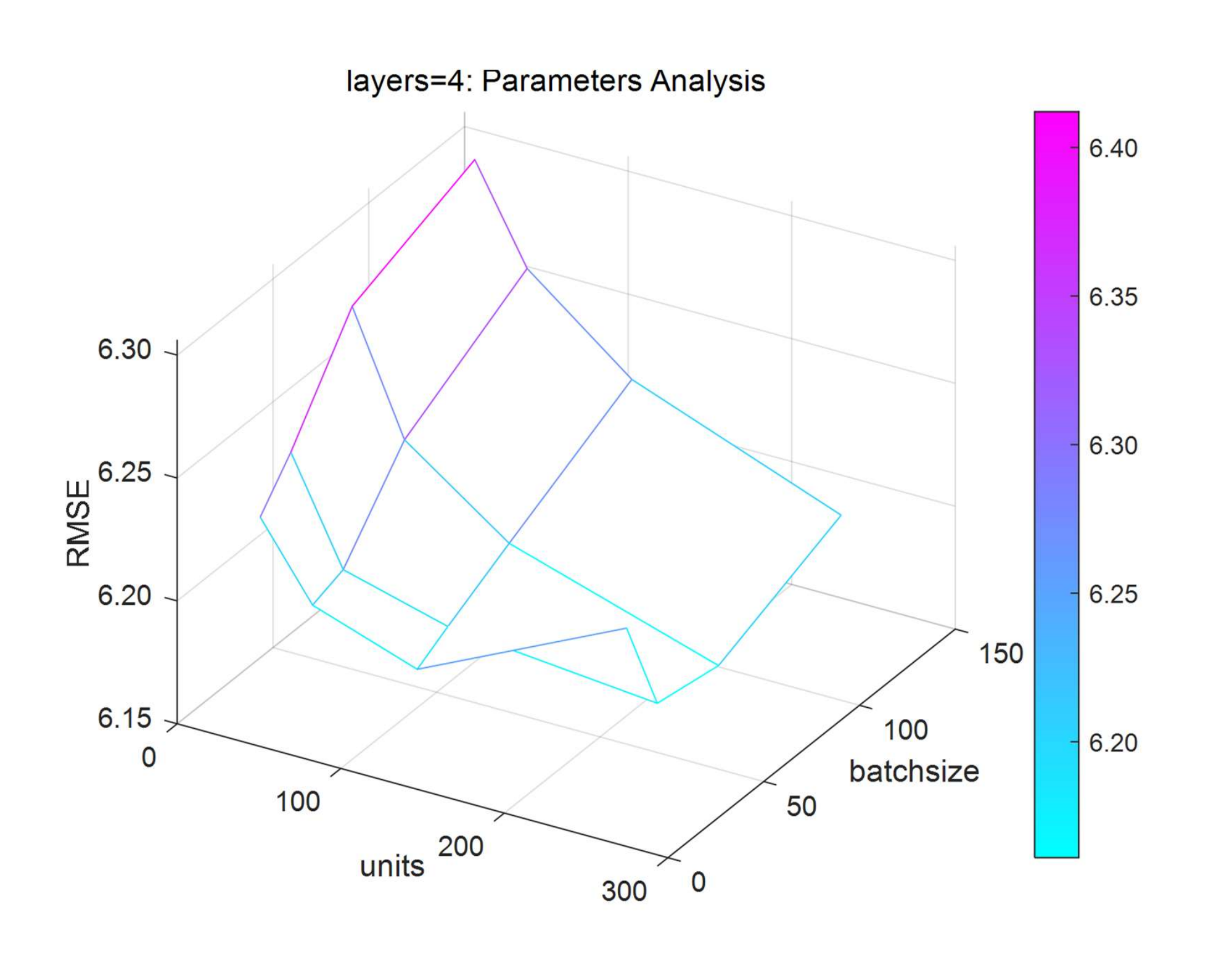}}
		\centerline{(d)}
	\end{minipage} 
	\caption{Effects of different paremeters  (a) MAE value of layers=3; (b) MAE value of layers=4; (c) RMSE value of layers=3; (d) RMSE value of layers=4.}
	\label{FIG:8}
\end{figure}

\begin{figure}[htb]
	\centering
	\includegraphics[scale=.48]{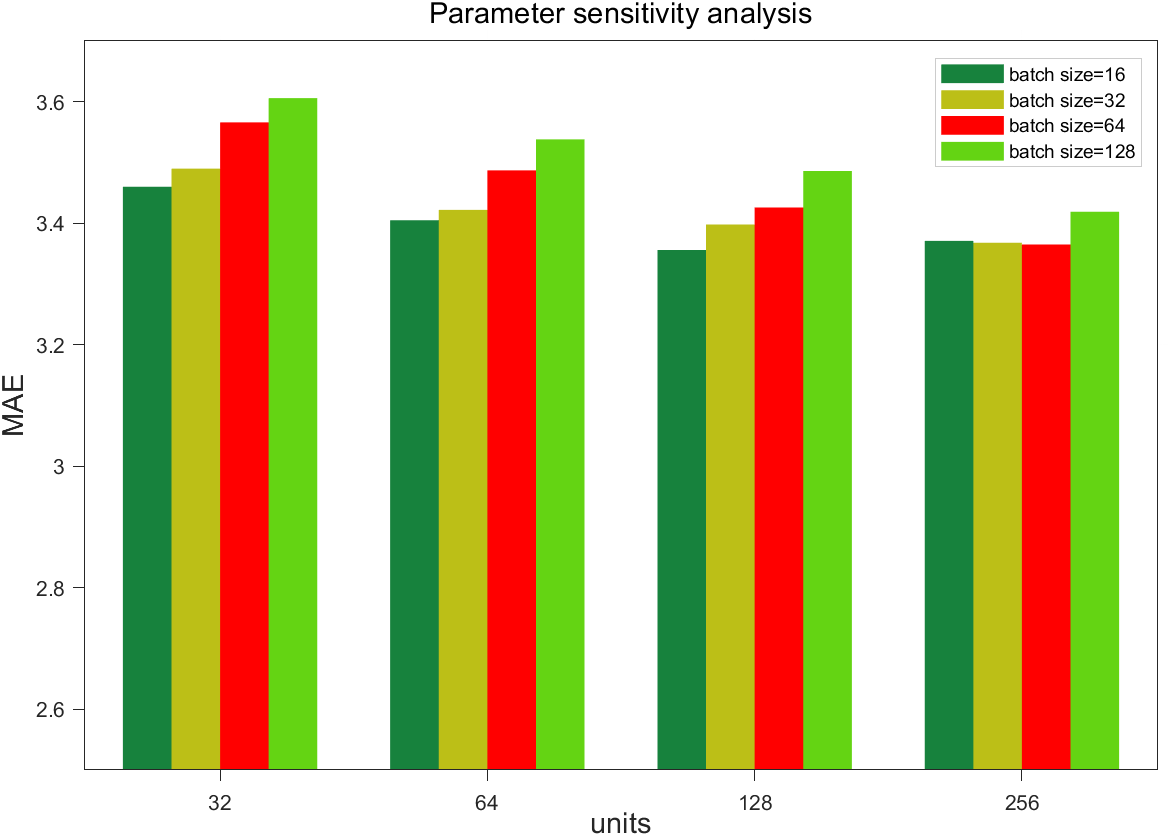}
	\caption{Parameter sensitivity analysis of layers=4.}
	\label{FIG:9}
\end{figure}

In the propsed model, the cosine annealing learning rate adjustment (CALRA) \citep{RN52} is used to improve the training performance. The period cycle number of CALRA is 4. The maximum initial learning rate and the minimum initial learning rate are 2.4e-5 and 1.5e-5, respectively. In addition, we obtain the performance with different parameters using grid search to configure an optimal combination \citep{liu2020image}, and the effects of different parameters are shown in Figure. \ref{FIG:8}. The batch size refers to the number of samples traversed before calculating the loss function. A smaller batch size may lead to a better model performance, but will increase computational cost. The similar effect can also be observed when the number of units increases. However, there is no absolute negative correlation among batch size, units and model performance as Figure \ref{FIG:9} shows. Therefore, we finally set the batch size and the numbers of the units to 64 and 256, respectively, after grid search. The setting of parameters in each layer is shown in Table \ref{tbl1}. Further analysis of the grid settings for each parameter and margin effect on prediction performance with the proposed model are described in detail in Appendix.

\begin{table}[width=.9\linewidth,cols=3,pos=h]
	\caption{Settings of each layer of the model.}\label{tbl1}
	\begin{tabular*}{\tblwidth}{@{} CCC@{} }
		\toprule
		Number & Name & Settings  \\
		\midrule
		1&	graph\_convolution\_1&	units=128,steps=1 \\
		2&	lstm\_1  & units=256,bias\_regularizer=l2(0.01)\\
		3&	lstm\_2  & units = 256,bias\_regularizer=l2(0.01)\\
		4&  fully-connected\_1  & units = 12\\
		\bottomrule
	\end{tabular*}
\end{table}

We also conduct grid search to obtain the best parameters for the baseline models. In LSTM model, the batch size is set to 32 and the number of the units is set to 256 for both LSTM layers, and finally a fully connected layer of 12 units is connected to yeild the output. In STGCN model, the number of the channels of three layers in ST-Conv block are set to 256, 64, 256 respectively; the graph convolution kernel size and temporal convolution kernel size are both set to 3. In addition, the learning rate is set to the same value as the proposed model, and batch size is set to 32. In DCRNN model, the maximum steps of random walks, i.e. $K$, is set to 1, the nubmers of the units in the convolution layer and in the two GRU layers are set to 64 and 128, respectively, and the batch size is set to 32; the decaying learning rate instead degrades the performance of this model, so we employ a fixed learning rate with a value of $3e-5$. In XGBoost, best results are achieved when the maximum depth is set to 6 and the subsample is set to 0.9. The Radial Basis Function (RBF) kernel and the penalty term $C=1$ are applied in SVR. The default parameters are used for LR. 
Only a limited number of parameter values can be examined using the grid search strategy, but it is clear from these results that the effect of the parameters on the performance is much less than the improvement brought by model structure. Therefore, we apply the above parameters to obtain the final results for comparison. The next subsection shows the results of the comparison.

\subsection{Results}

\subsubsection{Prediction results}

\begin{table}[width=.9\linewidth,cols=8,pos=h]
	\caption{Performance comparison between the proposed model and the selected baseline models on OpenITS dataset.}\label{tbl2}
	\begin{tabular*}{\tblwidth}{@{} CCCCCCCCC@{} }
		\toprule
		&LR & XGBoost & SVR  & LSTM  &  STGCN   &  DCRNN &  Loc-GCLSTM\\
		\midrule
		RMSE&12.325&11.547&11.287&9.463&9.982&8.176&\bf{7.768}\\
		MAE&9.173&8.626&8.389&7.095&7.497&6.181&\bf{5.879}\\
		MAPE(\%)&28.195&24.252&23.405&20.275&21.985&18.214&\bf{17.208}\\
		MdAE&7.150&6.794&6.507&5.452&5.808&4.814&\bf{4.598}\\
		MdAPE(\%)&28.205&24.133&23.323&20.239&22.083&18.177&\bf{17.198}\\
		\bottomrule
	\end{tabular*}
\end{table}

Table \ref{tbl2} shows the accuracy comparison between those models in the five evaluation criteria on OpenITS dataset. Loc-GCLSTM achieves the best results according to all criteria. It is worth noting that the spatial-temporal model STGCN performs worse than LSTM, due to the lack of LSTM for long-term dependency analysis. When compared with the LSTM, Loc-GCLSTM improves 17.91\% in RMSE, 17.14\% in MAE, and 1.51\% in MAPE. When compared with the DCRNN, Loc-GCLSTM improves 4.99\% in RMSE, 4.89\% in MAE, and 5.52\% in MAPE.

\begin{table}[width=.9\linewidth,cols=8,pos=h]
	\caption{Performance comparison between the proposed model and the selected baseline models on METR-LA dataset.}\label{tbl3}
	\begin{tabular*}{\tblwidth}{@{} CCCCCCCCC@{} }
		\toprule
		&LR & XGBoost & SVR  & LSTM  &  STGCN   &  DCRNN  &  Loc-GCLSTM\\
		\midrule
		RMSE&7.073&7.006&7.138&6.664&7.392&6.736&\bf{6.161}\\
		MAE&3.830&3.585&3.685&3.589&4.297&3.775&\bf{3.365}\\
		MAPE(\%)&10.215&9.548&9.666&9.968&12.087&10.467&\bf{9.104}\\
		MdAE&2.149&1.834&2.074&1.947&2.526&2.151&\bf{1.898}\\
		MdAPE(\%)&8.471&8.086&8.031&7.925&10.120&8.607&\bf{7.548}\\
		\bottomrule
	\end{tabular*}
\end{table}

Table \ref{tbl3} shows the accuracy comparison between those models in the five evaluation criteria on METR-LA dataset. Again, Loc-GCLSTM outperforms other methods according to all criteria. Since the task is to predict the traffic flow of the next twelve five-minute time intervals, the LSTM model also performs well on this dataset, which fully demonstrates the excellence of LSTM in performing multiple sequence prediction tasks. Loc-GCLSTM improves 7.54\% in RMSE, 6.08\% in MAE, and 8.67\% in MAPE compared with LSTM, and improves 8.54\% in RMSE, 10.86\% in MAE, and 13.02\% in MAPE compared with DCRNN. The improvement of the proposed model over other selected models in each evaluation criteria can also be obesevsed in the table.

\begin{figure}[htb]
	\centering
	\begin{minipage}[]{0.45\linewidth}
		\centerline{\includegraphics[scale=.35]{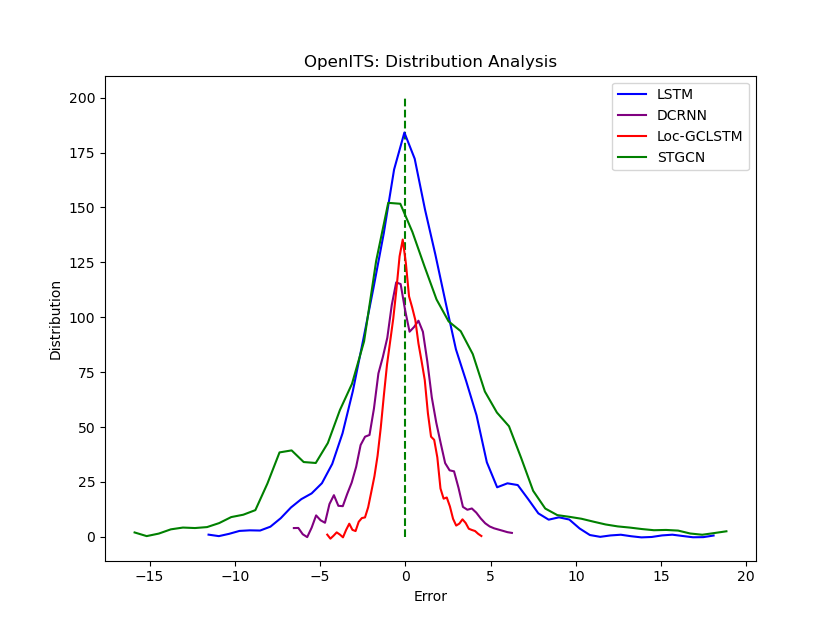}}
		\centerline{(a)}
	\end{minipage}  
	\begin{minipage}[]{0.45\linewidth}
		\centerline{\includegraphics[scale=.35]{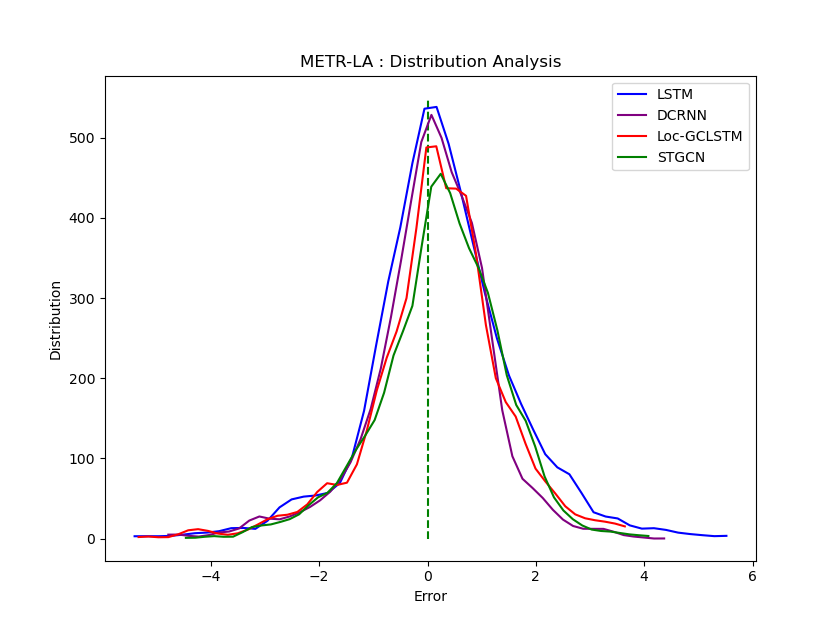}}
		\centerline{(b)}
	\end{minipage} 
	\caption{Comparison between the distribution of prediction error on (a) OpenITS dataset; (b) METR-LA dataset.}
	\label{FIG:10}
\end{figure}

Figure \ref{FIG:10} shows the distribution of the prediction error of LSTM, DCRNN, and Loc-GCLSTM on OpenITS dataset and METR-LA dataset, where X-axis is the error between the predicted result and the ground truth and Y-axis is the distribution of the error. It is obvious that the distribution of Loc-GCLSTM is denser near $0$, indicating that the Loc-GCLSTM obtain more accurate results in most cases. And it is also denser along X-axis, showing that Loc-GCLSTM is more robust in prediction accuracy.

\subsubsection{Convergence Rate}
\begin{figure}[htb]
	\centering
	\begin{minipage}[]{0.45\linewidth}
		\centerline{\includegraphics[scale=.5]{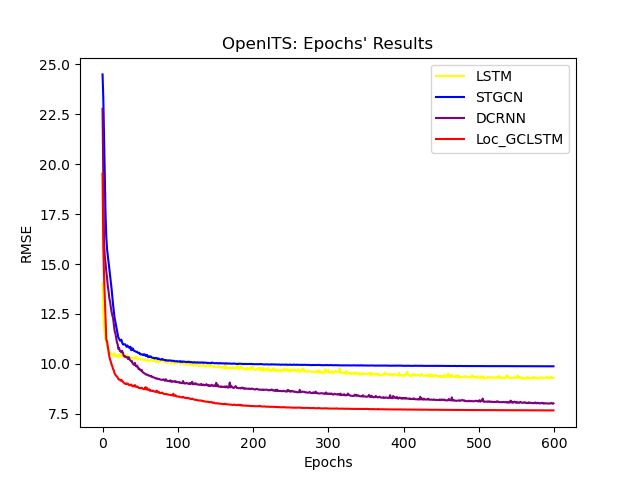}}
		\centerline{(a)}
	\end{minipage} 
	\begin{minipage}[]{0.45\linewidth}
		\centerline{\includegraphics[scale=.5]{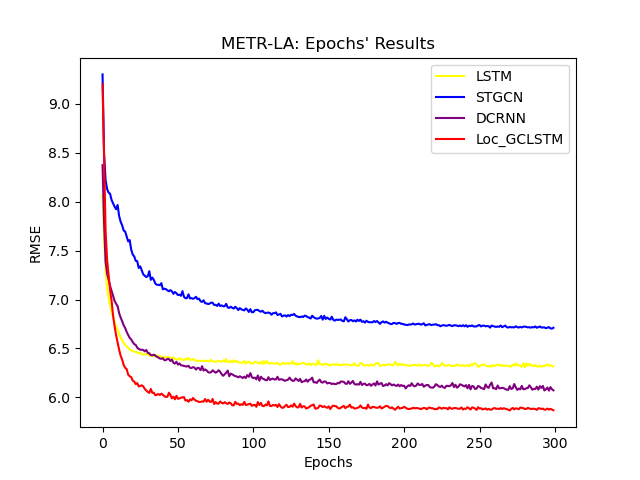}}
		\centerline{(b)}
	\end{minipage}  
	\caption{Comparison of convergence rate on (a) OpenITS dataset and (b) METR-LA dataset.}
	\label{FIG:11}
\end{figure}

The predicted results of LSTM, STGCN, DCRNN, Loc-GCLSTM on the test dataset are recorded after each epoch during the training process. RMSE is used to evaluate their convergence rate. Figure \ref{FIG:11} shows the evaluation results, where X-axis records the number of epochs and Y-axis the RMSE score, the yellow line coresponds to LSTM, the blue line corresponds to STGCN, the purple line corresponds to DCRNN, and the red one corresponds to Loc-GCLSTM. It is obvious that Loc-GCLSTM has the fastest convergence rate and the most stable and highest prediction accuracy on both OpenITS and METR-LA datasets.

\subsubsection{Prediction-Truth Comparison}
\begin{figure}[p]
	\begin{minipage}[]{0.45\linewidth}
		\centerline{\includegraphics[scale=.3]{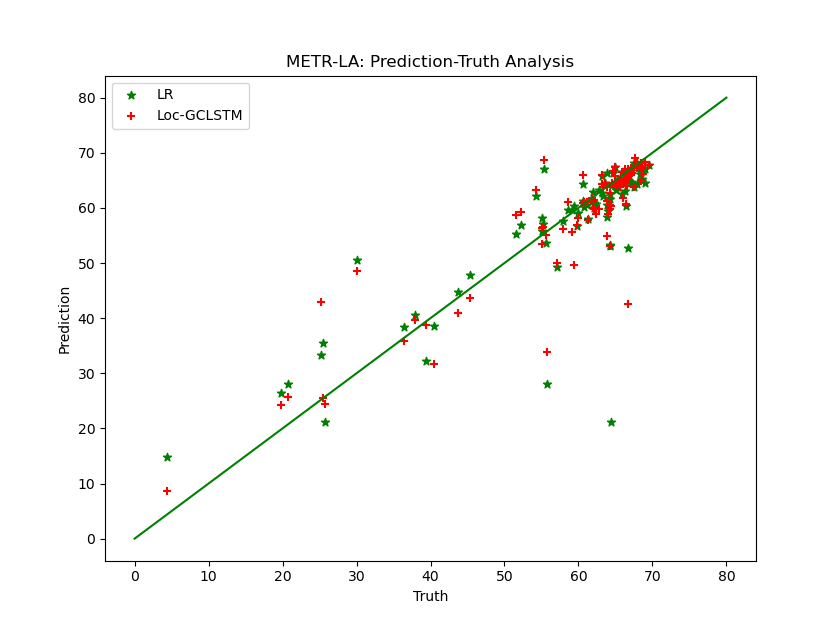}}
	\end{minipage}
	\hfill
	\begin{minipage}[]{0.45\linewidth}
		\centerline{\includegraphics[scale=.3]{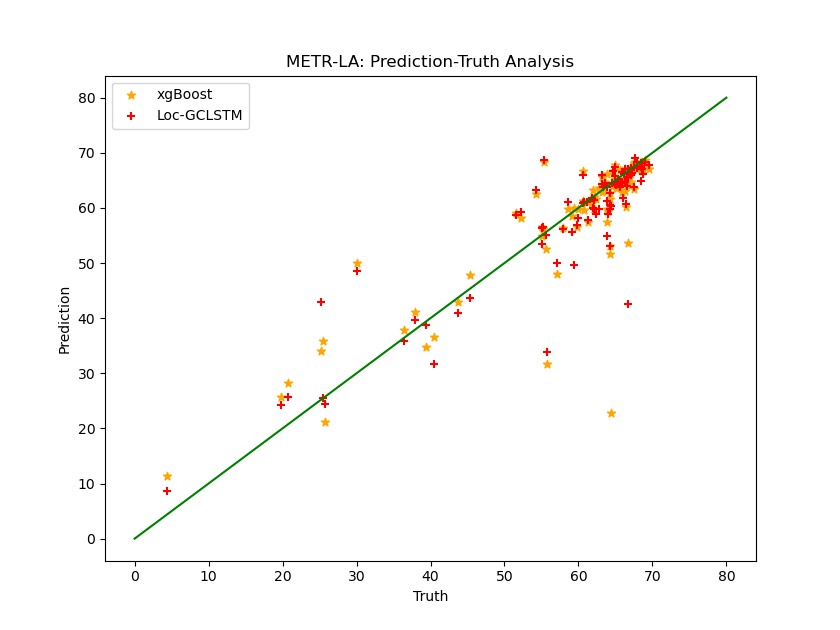}}
	\end{minipage}
	\vfill
	\begin{minipage}[]{0.25\linewidth}
		\centerline{\includegraphics[scale=.3]{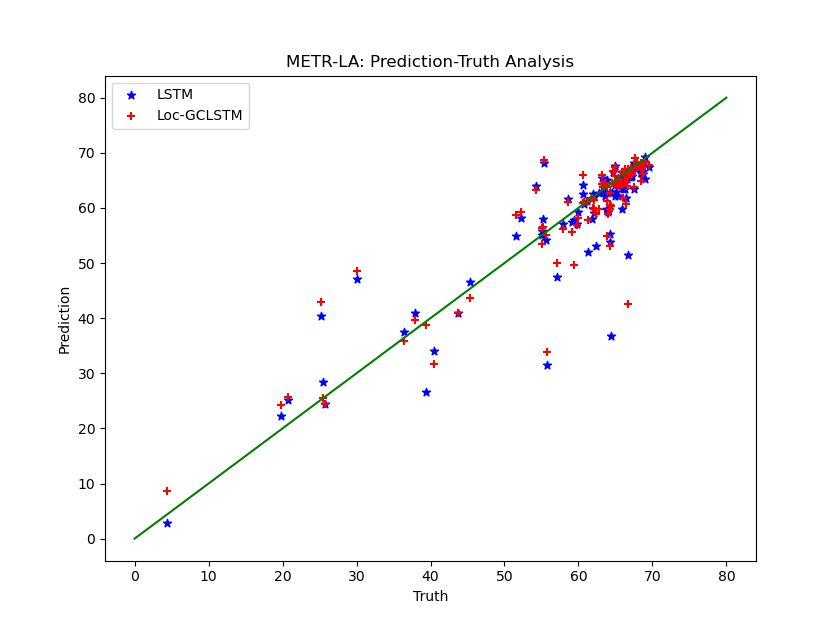}}
	\end{minipage}
	\hfill
	\begin{minipage}[]{0.25\linewidth}
		\centerline{\includegraphics[scale=.3]{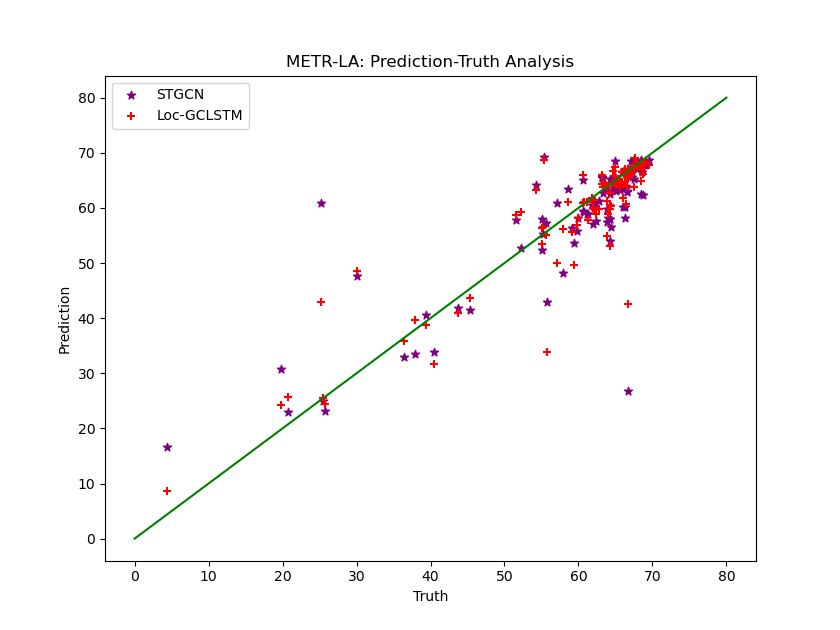}}
	\end{minipage}
	\hfill
	\begin{minipage}[]{0.25\linewidth}
		\centerline{\includegraphics[scale=.3]{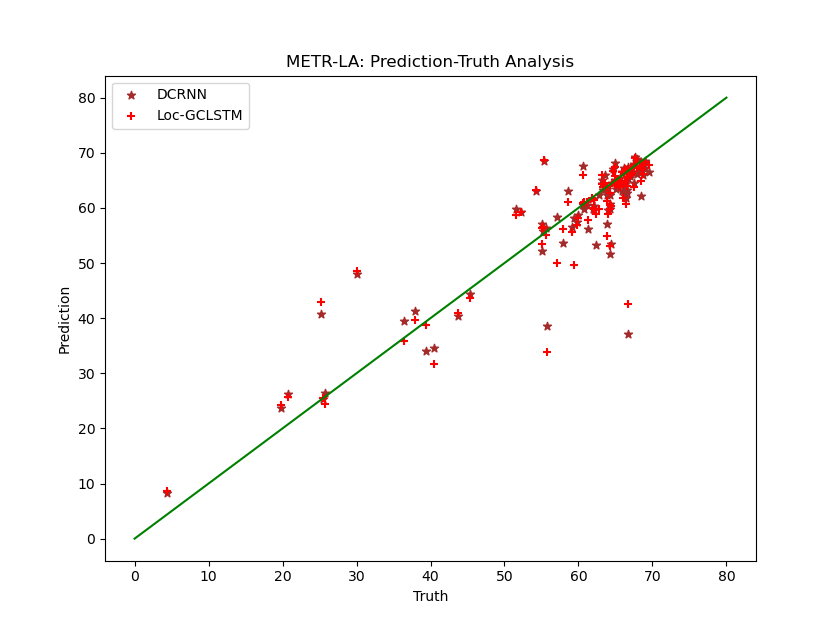}}
	\end{minipage}
	\vfill
	\centerline{(b)}
	\caption{Random-sampled comparison between the prediction results and the ground truth on (a) OpenITS dataset and (b) METR-LA dataset.}
	\label{FIG:12}
\end{figure}

Figure \ref{FIG:12} shows the comparison between the prediction result and the ground truth on both OpenITS and METR-LA datasets, where X-axis is the value of the ground truth and Y-axis is the value of the prediction result. In each picture, there are 100 random-sampled points of Loc-GCLSTM and a comparison model, respectively. It can be seen that sample points of Loc-GCLSTM, compared with other models, are denser around the line $y=x$, indicating higher accuracies.

\begin{figure}[p]
	\centering
	\begin{minipage}[]{0.25\linewidth}
		\centerline{\includegraphics[scale=.3]{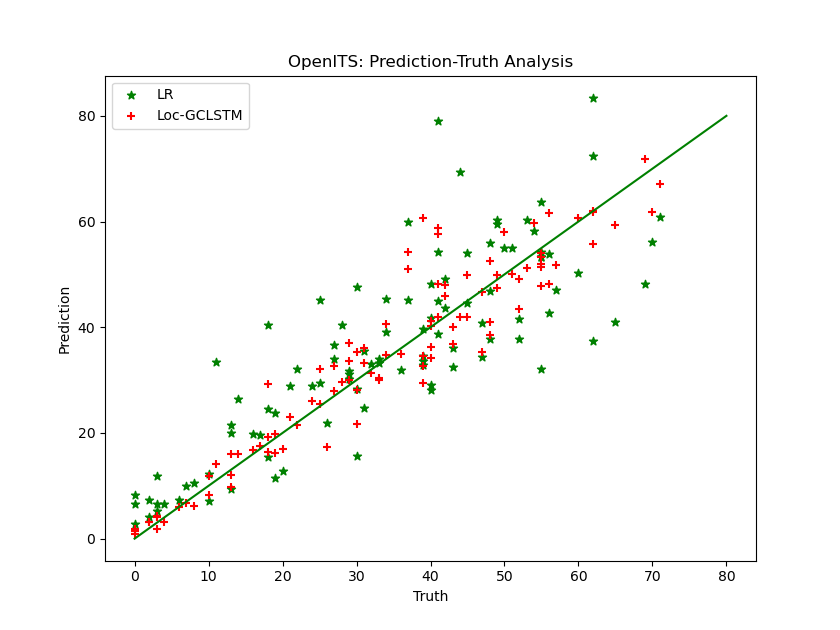}}
	\end{minipage}
	\hfill
	\begin{minipage}[]{0.25\linewidth}
		\centerline{\includegraphics[scale=.3]{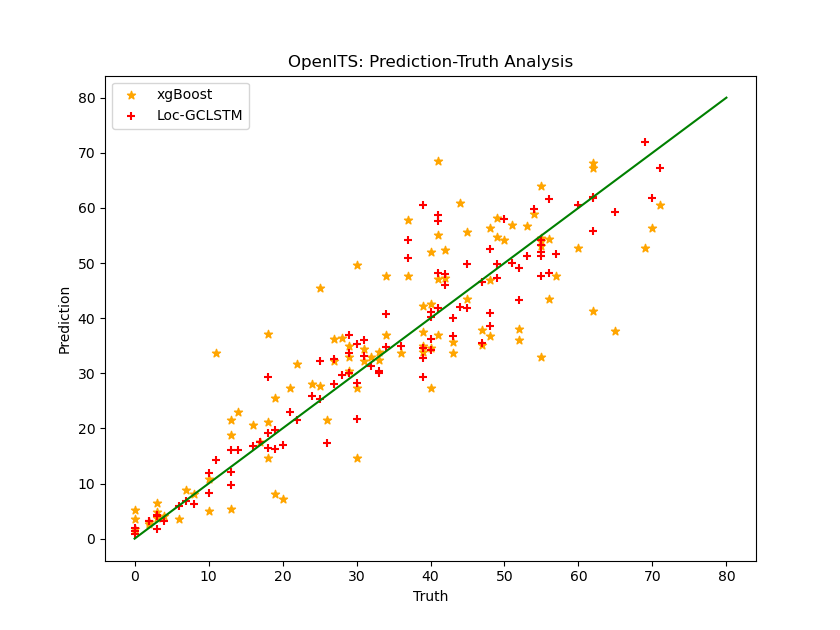}}
	\end{minipage}
	\hfill
	\begin{minipage}[]{0.25\linewidth}
		\centerline{\includegraphics[scale=.3]{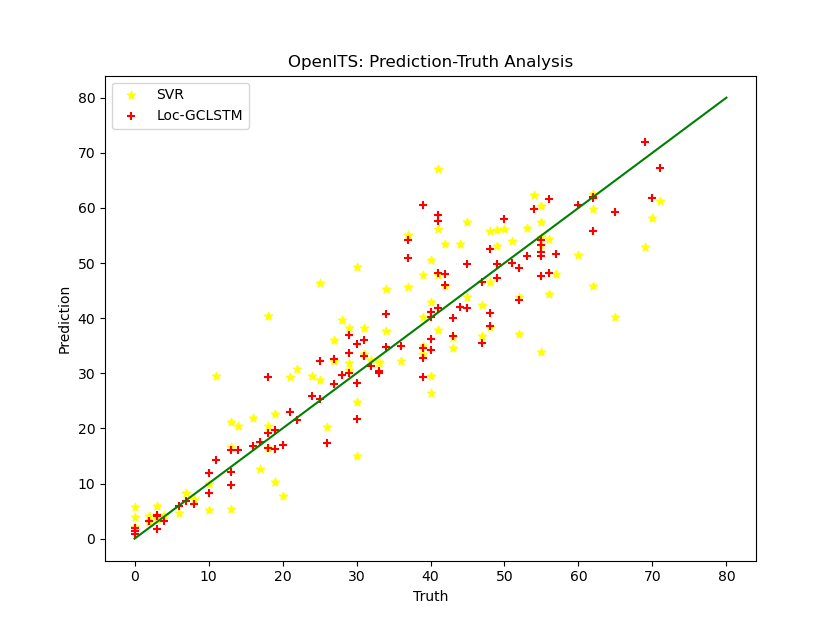}}
	\end{minipage}
	\vfill
	\begin{minipage}[]{0.25\linewidth}
		\centerline{\includegraphics[scale=.3]{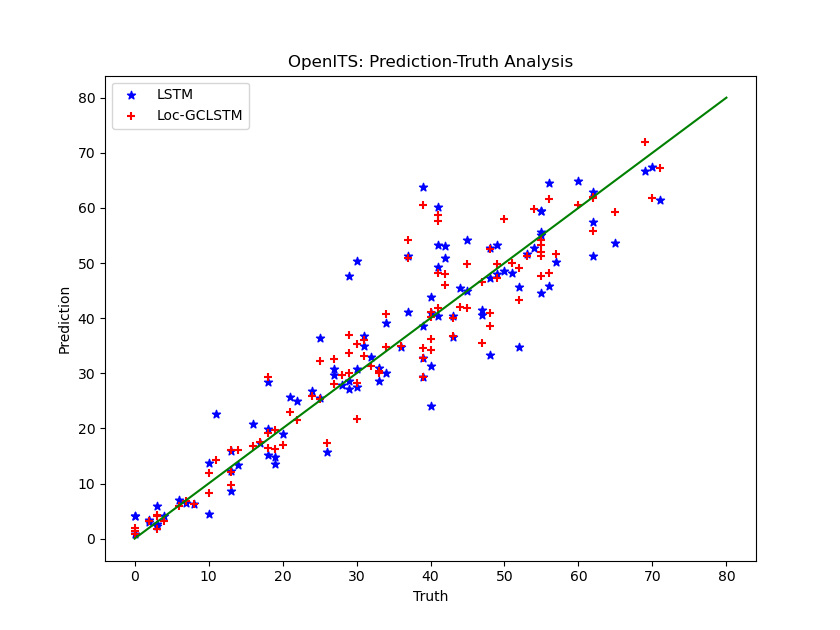}}
	\end{minipage}
	\hfill
	\begin{minipage}[]{0.25\linewidth}
		\centerline{\includegraphics[scale=.3]{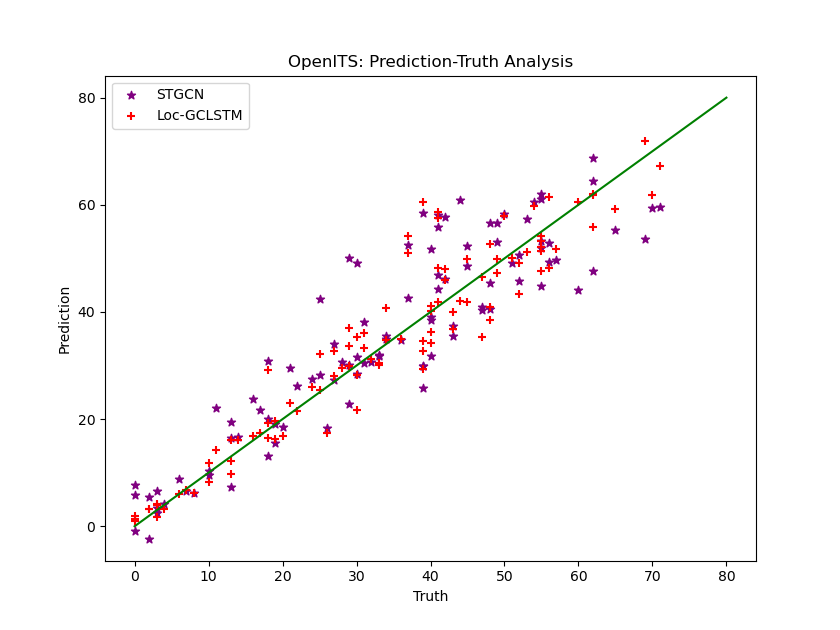}}
	\end{minipage}
	\hfill
	\begin{minipage}[]{0.25\linewidth}
		\centerline{\includegraphics[scale=.3]{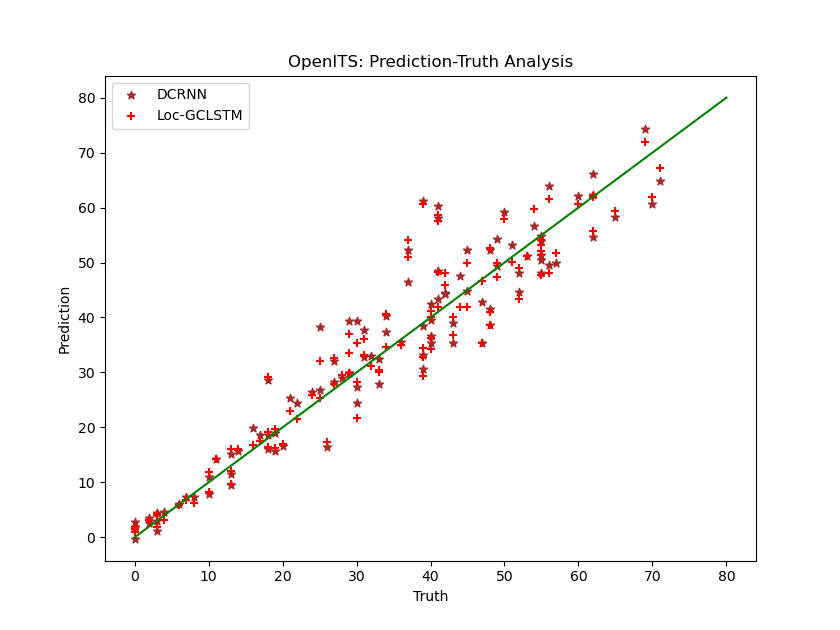}}
	\end{minipage}
	\centerline{(a)}
\end{figure}

\subsubsection{Predicted Results and Ground Truth}

\begin{figure}[htp]
	\centering
	\includegraphics[scale=.45]{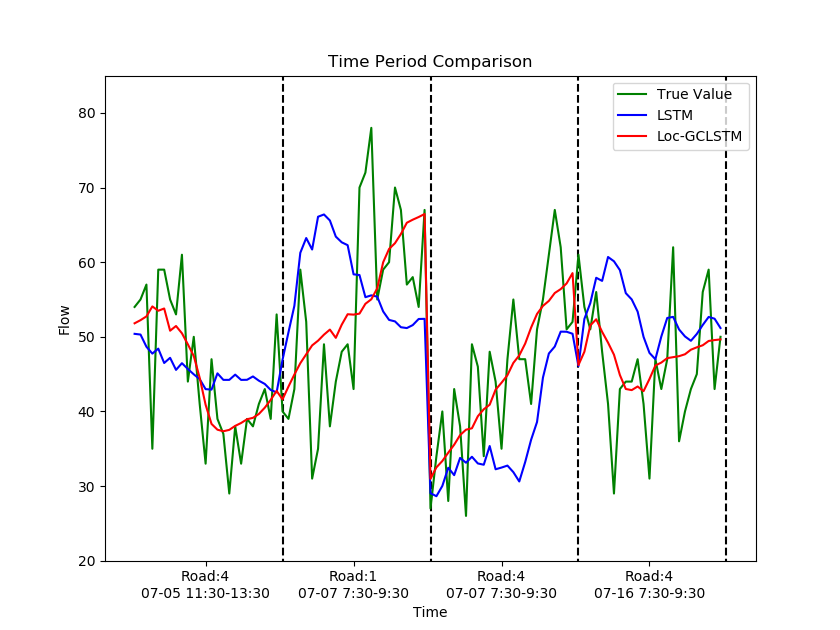}
	\caption{Comparison between predicted results and ground truth on OpenITS dataset.}
	\label{FIG:13}
\end{figure}

Under the evaluation of OpenITS dataset, the 1st and 4th observation node's 07:30-09:30, 11:30-13:30 on 5th July, 7th July, and 16th July, totally 4 time-slots' prediction results of LSTM and Loc-GCLSTM are chosen to be compared with the ground truth. As shown in Figure \ref{FIG:13}, the green lines are the ground truth; the blue lines are the predicted results of LSTM; the red lines are the prediction results of Loc-GCLSTM. It's easy to see that the prediction results of Loc-GCLSTM are closer to the ground truth.

\subsubsection{Predicted Accuracy on Each Road}

\begin{figure}[htp]
	\begin{minipage}[]{0.25\linewidth}
		\centerline{\includegraphics[scale=.4]{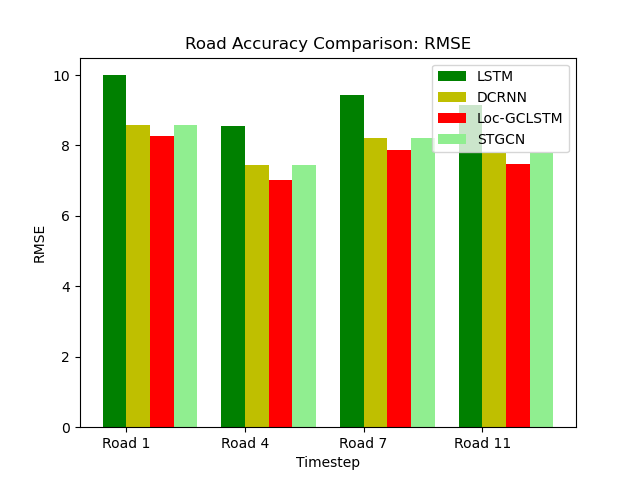}}
		\centerline{(a)}
	\end{minipage}
	\hfill
	\begin{minipage}[]{0.25\linewidth}
		\centerline{\includegraphics[scale=.4]{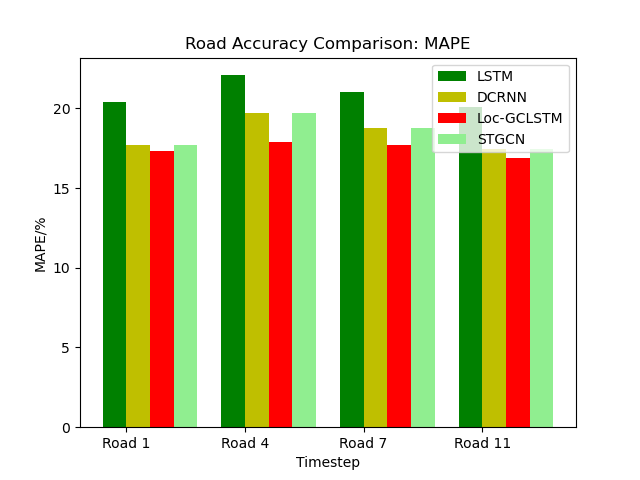}}
		\centerline{(b)}
	\end{minipage}
	\hfill
	\begin{minipage}[]{0.25\linewidth}
		\centerline{\includegraphics[scale=.4]{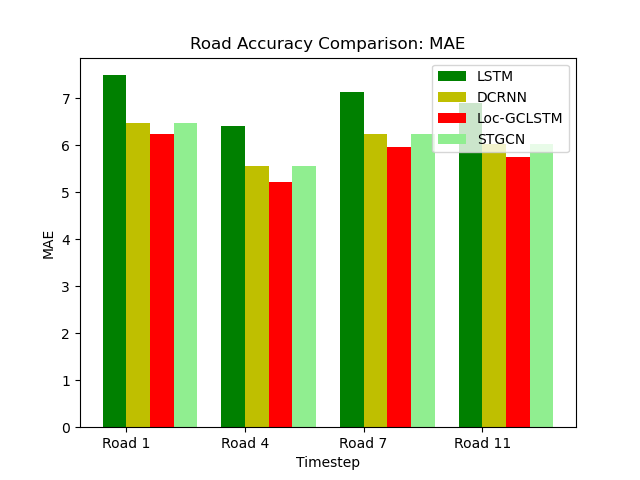}}
		\centerline{(c)}
	\end{minipage}
	
	\caption{Comparison of prediction results on each road among Loc-GCLSTM, DCRNN,  LSTM, and STGCN. (a) Comparison of RMSE; (b) Comparison of MAPE; (c) Comparison of MAE.}
	\label{FIG:14}
\end{figure}

As shown in Figure \ref{FIG:14}, Loc-GCLSTM is compared with normal LSTM, DCRNN and STGCN on each road's prediction accuracy under the evaluation of RMSE, MAPE, and MAE, under the evaluation of OpenITS dataset. On every selected test road, Loc-GCLSTM performs a prominent better prediction ability than LSTM and DCRNN. The accuracy improvements varied not much between each road section, meaning the Loc-GCLSTM can usually conduct a more accurate prediction in most real-life roads.

\subsubsection{Location Module Analysis}
\begin{figure}[htp]
	\centering
	\begin{minipage}[]{0.45\linewidth}
	\centerline{\includegraphics[scale=.4]{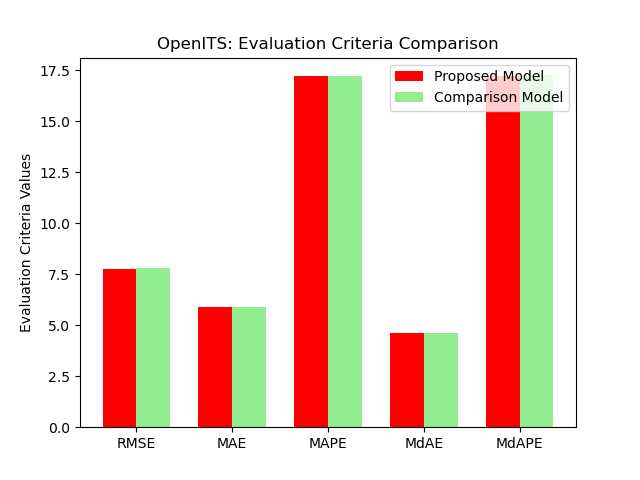}}
	\centerline{(a)}
	\end{minipage} 
	\begin{minipage}[]{0.45\linewidth}
	\centerline{\includegraphics[scale=.4]{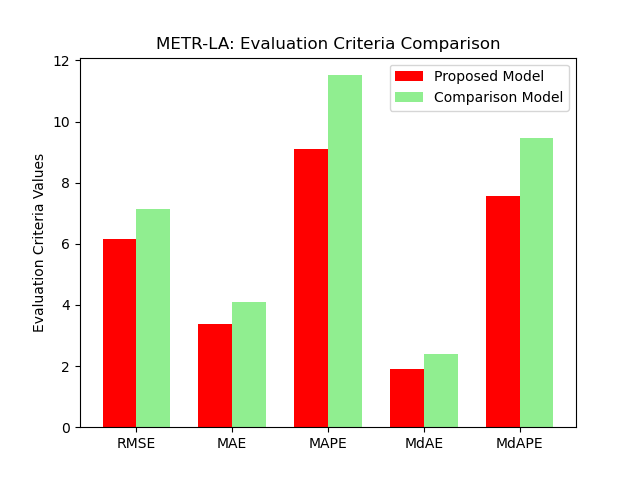}}
	\centerline{(b)}
	\end{minipage}  
	\caption{Effect of the location module on (a) OpenITS dataset and (b) METR-LA dataset.}
\label{FIG:15}
\end{figure}

To further analyze the effect of the location module on the performance of the proposed model in traffic flow prediction, we construct a comparison model without a location-based learning component for validation, in which all parameters are the same with the proposed model except for the GCN layer. As shown in Figure \ref{FIG:15}, the location module does not have a significant impact on the model performance on OpenITS dataset. However, on the META-LA dataset, the proposed location based learning component has a significant impact on the model performance in traffic flow prediction. This is possibly because that the key of our model to improve performance is to capture the different impact weights between roads. However, the number of roads in OpenITS dataset is small and the amount of data is so small that the proposed component does not fully demonstrate its performance. But when it comes to the META-LA dataset, which has a sufficient number of roads and sample size, the location module can fully exploit the potential information between road nodes, thus significantly improving the accuracy of traffic flow prediction.

\section{Conclusion}
In this paper, a new short-term traffic flow prediction model named Loc-GCLSTM is proposed. The model combines Location-GCN with LSTM to conduct prediction from both temporal and spatial perspectives. Location-GCN enables the graph convolution mechanism to learn different influence weights among nodes dynamically during the training, using the absolute value of an added trainable matrix. Moreover, the Trigonometric Function Encoding is used to preprocess the time point data, making the short-term input sequence able to convey periodical patterns into the network model. 

Experimental results show that the proposed Loc-GCLSTM model can achieve the best prediction performance compared with various respective models, on both OpenITS and METR-LA datasets. It also has good robustness compared with other models, which means a better application potential in real-life. The convergence rate of our model is also faster than other compared models, meaning less requirement for computing resources. Moreover, our main contributions like Location-GCN are not conflicting with those advanced models such as DCRNN and GC-Seq2Seq, they can also be used in those models.

However, the Location-GCN used in our paper only considers the difference among different road sections'influences. It's still not able to learn how the influence weights among different nodes change over time. And since Location-GCN is still based on the traditional GCN mechanism, the node sections of the road network can not be changed in both the training and testing process, limiting the convenience and flexibility of the model in real-life usage. All those drawbacks worth further study.

\printcredits

\section*{Declaration of Competing Interest}
The authors declare that they have no known competing financial interests or personal relationships that could have appeared to influence the work reported in this paper.

\section*{Acknowledgements}
The authors wish to thank the anonymous reviewers and the Editors of the journal for their constructive comments which improve the readability of the paper.

\bibliographystyle{cas-model2-names}

\bibliography{refs}

\clearpage
\section*{Appendix} 
The appendix describes further analysis of the grid settings for each parameter and effects on prediction performance with the proposed model. To optimize the model for more accurate traffic flow prediction results, the parameters of the model need to be selected, often known as parameter tuning. There are several approaches for tuning. For our study, we choose the method of grid search, which is an exhaustive search of a subset of the hyperparameter space of the algorithm \citep{priyadarshini2021novel}, i.e., each possible combination of parameters is tried, and finally the optimal combination of parameters is selected by evaluating the performance of each combination according to the evaluation criteria mentioned in the previous section. 

\begin{table}[width=.9\linewidth,cols=3,pos=h]
	\caption{Grid search hyperparameters settings.}\label{tbl4}
	\begin{tabular*}{\tblwidth}{@{} CCC@{}}
		\toprule
		Number & Hyperparameters & Values  \\
		\midrule
		1&	batch size &16,32,64,128 \\
		2&	units & 32,64,128,256\\
		3&	layers  & 3,4\\
		\bottomrule
	\end{tabular*}
\end{table}

\begin{figure}[htbp]
	\centering
	\begin{minipage}[t]{0.45\linewidth}
		\centerline{\includegraphics[scale=.4]{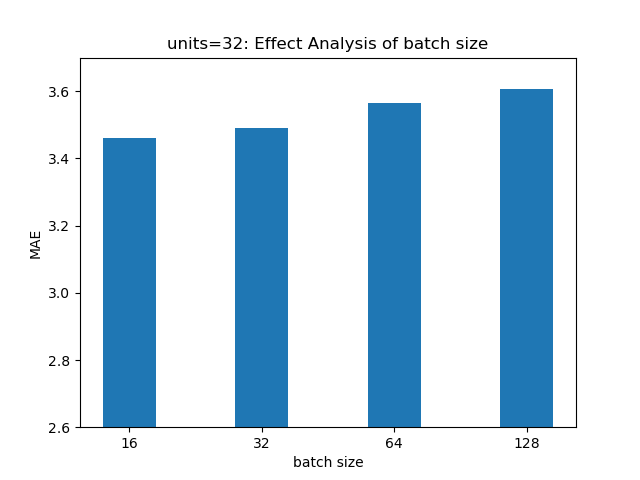}}
		\centerline{(a)}
	\end{minipage} 
	\hfill
	\begin{minipage}[t]{0.45\linewidth}
		\centerline{\includegraphics[scale=.4]{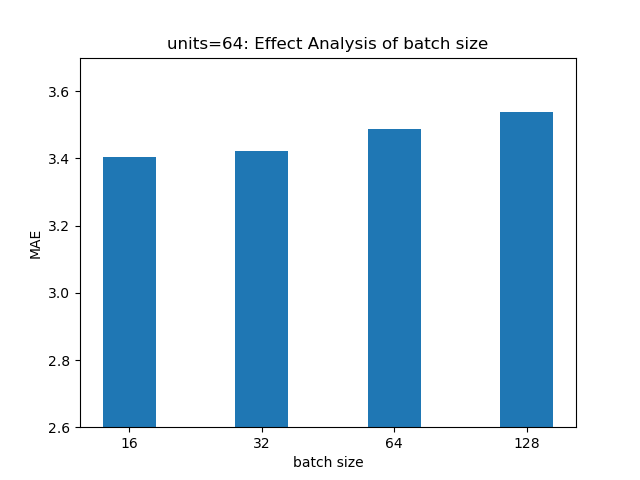}}
		\centerline{(b)}
	\end{minipage} 
	\hfill
	\begin{minipage}[t]{0.45\linewidth}
		\centerline{\includegraphics[scale=.4]{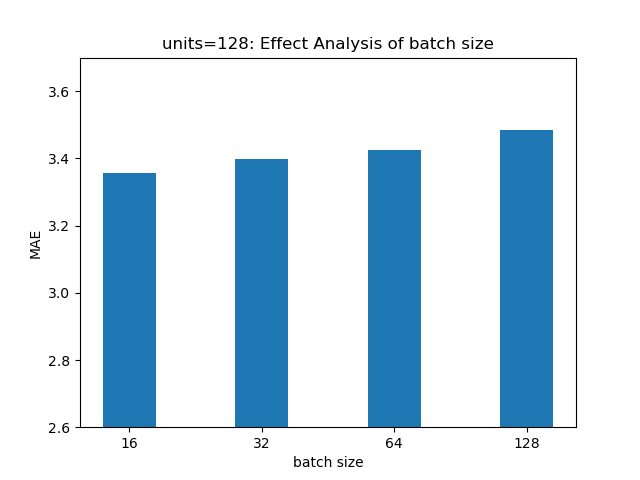}}
		\centerline{(c)}
	\end{minipage} 
	\hfill
	\begin{minipage}[t]{0.45\linewidth}
		\centerline{\includegraphics[scale=.4]{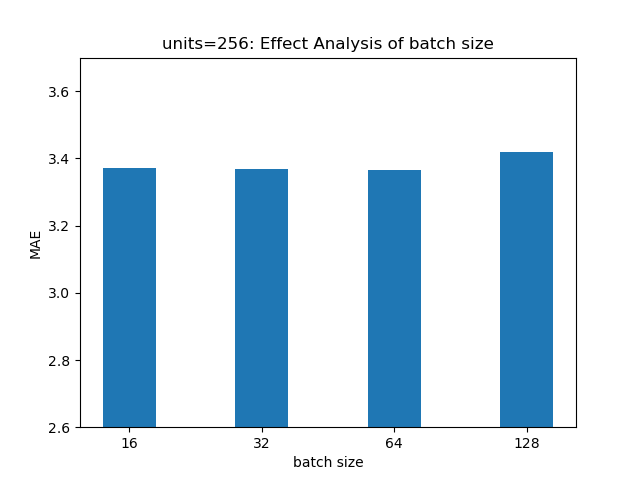}}
		\centerline{(d)}
	\end{minipage} 
	\caption{Effects of different units and batch size setting  (a) MAE value of different batch size when units=32; (b) MAE value of different batch size when units=64; (c) MAE value of different batch size when units=128; (d)MAE value of different batch size when units=256}
	\label{FIG:16}
\end{figure}

\begin{table}[width=.9\linewidth,cols=8,pos=t]
	\caption{Grid search parameter combination settings and performance evaluation scores when layers=4.}\label{tbl5}
	\begin{tabular*}{\tblwidth}{@{} CCCCCCCCC@{} }
		\toprule
		Layers=4 &
		Metrics &
		units=32 &
		units=64 &
		units=128 &
		units=256 \\
		\midrule
		batch size=16 &
		\begin{tabular}[c]{@{}l@{}}MSE\\ RMSE\\ MAE\\ MAPE\\ MdAE\\ MdAPE\end{tabular} &
		\begin{tabular}[c]{@{}l@{}}39.261\\ 6.260\\ 3.460\\ 9.293\\ 1.986\\ 7.732\end{tabular} &
		\begin{tabular}[c]{@{}l@{}}38.501\\ 6.200\\ 3.405\\ 9.180\\ 1.938\\ 7.603\end{tabular} &
		\begin{tabular}[c]{@{}l@{}}38.137\\ 6.171\\ 3.356\\ 9.082\\ 1.886\\ 7.526\end{tabular} &
		\begin{tabular}[c]{@{}l@{}}39.124\\ 6.251\\ 3.371\\ 9.092\\ 1.878\\ 7.553\end{tabular} \\
		batch size=32 &
		\begin{tabular}[c]{@{}l@{}}MSE\\ RMSE\\ MAE\\ MAPE\\ MdAE\\ MdAPE\end{tabular} &
		\begin{tabular}[c]{@{}l@{}}39.661\\ 6.293\\ 3.490\\ 9.358\\ 2.022\\ 7.769\end{tabular} &
		\begin{tabular}[c]{@{}l@{}}38.607\\ 6.209\\ 3.422\\ 9.217\\ 1.956\\ 7.640\end{tabular} &
		\begin{tabular}[c]{@{}l@{}}38.326\\ 6.186\\ 3.398\\ 9.153\\ 1.934\\ 7.589\end{tabular} &
		\begin{tabular}[c]{@{}l@{}}38.122\\ 6.170\\ 3.368\\ 9.047\\ 1.883\\ 7.480\end{tabular} \\
		batch size=64 &
		\begin{tabular}[c]{@{}l@{}}MSE\\ RMSE\\ MAE\\ MAPE\\ MdAE\\ MdAPE\end{tabular} &
		\begin{tabular}[c]{@{}l@{}}40.684\\ 6.372\\ 3.566\\ 9.570\\ 2.105\\ 7.954\end{tabular} &
		\begin{tabular}[c]{@{}l@{}}39.447\\ 6.275\\ 3.487\\ 9.382\\ 2.016\\ 7.776\end{tabular} &
		\begin{tabular}[c]{@{}l@{}}38.671\\ 6.231\\ 3.426\\ 9.214\\ 1.958\\ 7.641\end{tabular} &
		\begin{tabular}[c]{@{}l@{}}38.015\\ 6.161\\ 3.365\\ 9.104\\ 1.898\\ 7.548\end{tabular} \\
		batch size=128 &
		\begin{tabular}[c]{@{}l@{}}MSE\\ RMSE\\ MAE\\ MAPE\\ MdAE\\ MdAPE\end{tabular} &
		\begin{tabular}[c]{@{}l@{}}41.191\\ 6.412\\ 3.606\\ 9.721\\ 2.152\\ 8.093\end{tabular} &
		\begin{tabular}[c]{@{}l@{}}40.216\\ 6.335\\ 3.538\\ 9.495\\ 2.068\\ 7.904\end{tabular} &
		\begin{tabular}[c]{@{}l@{}}39.344\\ 6.268\\ 3.486\\ 9.355\\ 2.022\\ 7.791\end{tabular} &
		\begin{tabular}[c]{@{}l@{}}38.549\\ 6.204\\ 3.419\\ 9.249\\ 1.955\\ 7.671\end{tabular} \\
		\bottomrule
	\end{tabular*}
\end{table}

To obtain the optimal parameters combination of the proposed model, we conduct experiments on the same dataset as \citep{RN18}, i.e., METR-LA dataset and search left and right based on its grid settings when performing the grid search. Grid search hyperparameters considered in our study are dipicted in Table \ref{tbl4}.The specific grid for each model parameter combination and performance evaluation are shown in Table \ref{tbl5}. Figure \ref{FIG:16} shows the MAE values of each parameter setting when layers is set to 4 in the proposed model. The batch size refers to the number of samples traversed before calculating the loss function. We observers that smaller batch size enables the model to have higher prediction accuracy at the expense of increasing computational cost when units are set to 32, 64, 128. Similar effect is observed when the number of units increase. But when units are set to 256, with the decrease of batch size, the error on the validation dataset first quickly decrease, and then slightly increase. Considering the sensitivity and margin effects of the above two parameters on the prediction accuracy and the impact on the computational cost, we finally set the batch size and the units to 64 and 256 in the proposed model, respectively.


\end{document}